\documentclass{article}
\usepackage{PRIMEarxiv}
\usepackage[utf8]{inputenc} 
\usepackage[T1]{fontenc}    
\usepackage{url}            
\usepackage{booktabs}       
\usepackage{amsfonts}       
\usepackage{nicefrac}       
\usepackage{microtype}      
\usepackage{lipsum}
\usepackage{fancyhdr}       
\usepackage{graphicx}       
\graphicspath{{media/}}     
\usepackage{xcolor}
\usepackage[acronym]{glossaries}
\usepackage{siunitx}
\usepackage{appendix}
\usepackage{multirow}
\usepackage{caption}
\usepackage{subcaption}
\usepackage{hyperref}
\usepackage{tabularray}
\usepackage{amssymb}
\usepackage{pifont}
\newcommand{\cmark}{\ding{51}}%
\newcommand{\xmark}{\ding{55}}%
\usepackage[font=small]{caption}
\usepackage[hang,flushmargin]{footmisc}
\usepackage[flushleft]{threeparttable}
\usepackage[
backend=biber,
style=ieee,
citestyle=numeric-comp,
sorting=none,
url=false,
doi=true,
isbn=false,
uniquename=false,
alldates=year,
clearlang=true,]{biblatex} 

\AtEveryBibitem{\clearlist{publisher}}
\AtEveryBibitem{\clearfield{note}}
\AtEveryBibitem{\clearlist{institution}}
\AtEveryBibitem{\clearlist{location}}
\AtEveryBibitem{\clearlist{language}}
\AtEveryBibitem{\clearname{editor}}

\usepackage{doi}

\DeclareSIUnit\mac{MAC}

\DeclareFieldFormat[article,unpublished,misc,preprint,inproceedings]{title}{``#1''}

\let\oldcite\cite
\renewcommand{\cite}[1]{{\mbox{\oldcite{#1}}}}

\title{TEXEL: A neuromorphic processor with on-chip learning for beyond-CMOS device integration}

\author{
  Hugh Greatorex\textsuperscript{1,2,*}\\
  \\\And
  Ole Richter\textsuperscript{1,2}\\
  \\\And
  Michele Mastella\textsuperscript{1,2}\\
  \\\And
  Madison Cotteret\textsuperscript{1,2,3}\\
  \\\AND
  Philipp Klein\textsuperscript{1,2}
  \\\And
  Maxime Fabre\textsuperscript{1,2,4}\\
  \\\And
  Arianna Rubino\textsuperscript{5}\\
  \\\And
  Willian Soares Girão\textsuperscript{1,2}\\
  \\\AND
  Junren Chen\textsuperscript{5}\\
  \\\And
  Martin Ziegler\textsuperscript{3}\\
  \\\And
  Laura Bégon-Lours\textsuperscript{6}\\
  \\\And
  Giacomo Indiveri\textsuperscript{5}\\
  \\\AND
  Elisabetta Chicca\textsuperscript{1,2,*}\\
}

\PassOptionsToPackage{acronym}{glossaries}

\glsdisablehyper

\newcommand*\myglsentry[1]{%
   \glsentrylong{#1}%
}

\newacronym[longplural={Frames per Second}]{fpsLabel}{FPS}{Frame per Second}
\newacronym[longplural={Tables of Contents}]{tocLabel}{TOC}{Table of Content}
\newacronym{act}{ACT}{Asynchronous Circuit Toolkit}
\newacronym{adc}{ADC}{Analog Digital Converter}
\newacronym{aer}{AER}{Address Event Representation}
\newacronym{afe}{AFE}{Analog FrontEnd}
\newacronym{ai}{AI}{Artificial Intelligence}
\newacronym{ampa}{AMPA}{$\alpha$-amino-3-hydroxy-5-methyl-4-isoxazolepropionic acid}
\newacronym{ams}{AMS}{Analog Mixed-Signal}
\newacronym{api}{API}{Application Programming Interface}
\newacronym{bap}{BAP}{back-propagating action potential}
\newacronym{bcall}{BCaLL}{Bistable Calcium-based Local Learning}
\newacronym{bd}{BD}{Bundled Data}
\newacronym{beol}{BEOL}{Back-End Of Line}
\newacronym{bjt}{BJT}{Bipolar Junction Transistor}
\newacronym{bnc}{BNC}{Bayonet-Neill-Concelman}
\newacronym{bptt}{BPTT}{Back Propagation Through Time}
\newacronym{bp}{BP}{Back Propagation}
\newacronym{camp}{cAMP}{cyclic Adenosine Mono Phosphate}
\newacronym{cam}{CAM}{Content Addressable Memory}
\newacronym{cco}{CCO}{Current Controlled Oscillator}
\newacronym{cc-by}{CC-BY}{Creative Commons Attribution}
\newacronym{chp}{CHP}{Communicating Hardware Processes}
\newacronym{cl}{CL}{Closed-Loop}
\newacronym{cm}{CM}{Continuum Mechanics}
\newacronym{cpu}{CPU}{Central Process Unit}
\newacronym{csv}{CSV}{Comma-Separated Values}
\newacronym{ctat}{CTAT}{Complementary To Absolute Temperature}
\newacronym{cuba}{CUBA}{Current Based}
\newacronym{dac}{DAC}{Digital to Analog Converter}
\newacronym{dcn}{DCN}{Dorsal Column Nuclei}
\newacronym{dc}{DC}{Direct Current}
\newacronym{dpi}{DPI}{Differential Pair Integrator}
\newacronym{dpss}{DPSS}{Dendritic Prediction of Somatic Spiking}
\newacronym{drg}{DRG}{Dorsal Root Ganglion}
\newacronym{dr}{DR}{Dual Rail}
\newacronym{dsnn}{D-SNN}{Deep Spiking Neural Network}
\newacronym{dsp}{DSP}{Digital Signal Processing}
\newacronym{ds}{DS}{Delay Sensitive}
\newacronym{dvs}{DVS}{Dynamic Vision Sensor}
\newacronym{eda}{EDA}{Electronic Design Automation}
\newacronym{elm}{ELM}{Extreme Learning Machine}
\newacronym{esn}{ESN}{Echo State Network}
\newacronym{fac}{FAC}{Facilitatory Trace}
\newacronym{fecap}{FeCap}{Ferroelectric Capacitor}
\newacronym{fet}{FET}{Field Effect Transistor}
\newacronym{fft}{FFT}{Fast Fourier Transform}
\newacronym{fifo}{FIFO}{First In First Out Register}
\newacronym{fi}{FI}{Frequency vs. Current}
\newacronym{fm}{FM}{Frequency Modulated}
\newacronym{fo}{FO}{First Order}
\newacronym{fpga}{FPGA}{Field Programmable Gate Array}
\newacronym{fsm}{FSM}{Finite State Machine}
\newacronym{ftj}{FTJ}{Ferroelectric Tunnel Junction}
\newacronym{ft}{FT}{Fourier transform}
\newacronym{gmm}{GMM}{Gaussian mixture models}
\newacronym{gpu}{GPU}{Graphic Process Unit}
\newacronym{hp}{HP}{High Pass Filter}
\newacronym{i2c}{I$^2$C}{Inter Integrated Circuit}
\newacronym{ic}{IC}{Integrated Circuit}
\newacronym{if}{IF}{Integrate and Fire}
\newacronym{imc}{IMC}{In-Memory Computing}
\newacronym{iot}{IoT}{Internet of Things}
\newacronym{io}{I/O}{Input and Output}
\newacronym{ip}{IP}{Intellectual Property Macro}
\newacronym{irh}{IRH}{Instantaneous Rate Histogram}
\newacronym{isi}{ISI}{Interspike Interval}
\newacronym{iwta}{I-WTA}{Inverted Winner-Take-All}
\newacronym{knn}{K-NN}{K-Nearest Neighbors}
\newacronym{lcadc}{LC-ADC}{Level Crossing Analog Digital Converter}
\newacronym{lda}{LDA}{Linear Discriminant Analysis}
\newacronym{lif}{LIF}{Leaky \myglsentry{if}}
\newacronym{llm}{LLM}{Large Language Model}
\newacronym{lpf}{LPF}{Low Pass Filter}
\newacronym{lstm}{LSTM}{Long Short-Term Memory}
\newacronym{ltp}{LTP}{Long Term Plasticity}
\newacronym{lut}{LUT}{Look Up Table}
\newacronym{lvds}{LVDS}{Low Voltage Differential Signalling}
\newacronym{mac}{MAC}{Multiply Accumulate}
\newacronym{mems}{MEMS}{Micro Electro-Mechanical System}
\newacronym{mim}{MIM}{Metal Insulator Metal}
\newacronym{mlm}{MLM}{Multi Layer Mask}
\newacronym{mlp}{MLP}{Multi Layer Perceptron}
\newacronym{ml}{ML}{Machine Learning}
\newacronym{mnist}{MNIST}{modified National Institute of Standards and Technology database}
\newacronym{mom}{MOM}{Metal Oxide Metal}
\newacronym{mos}{MOS}{Metal Oxide Semiconductor}
\newacronym{mtj}{MTJ}{Magnetic Tunnel Junction}
\newacronym{mvm}{MVM}{Matrix-Vector Multiplication}
\newacronym{nas}{NAS}{Neuromorphic Auditory Sensor}
\newacronym{nda}{NDA}{Non Disclosure Agreement}
\newacronym{nmda}{NMDA}{N-methyl-D-aspartate}
\newacronym{nni}{NNI}{Neural Network Intelligence}
\newacronym{nn}{NN}{Neural Network}
\newacronym{noc}{NoC}{Network on Chip}
\newacronym{nvm}{NVM}{Non-Volatile Memory}
\newacronym{oa}{OA}{OpenAccess}
\newacronym{opamp}{OPAMP}{Operational Amplifier}
\newacronym{ota}{OTA}{Operational Transconductance Amplifier}
\newacronym{pca}{PCA}{Principal Component Analysis}
\newacronym{pcb}{PCB}{Printed Circuit Board}
\newacronym{pcfb}{PCFB}{Pre-Charge Full Buffer}
\newacronym{pchb}{PCHB}{Pre-Charge Half Buffer}
\newacronym{pcm}{PCM}{Phase Change Material}
\newacronym{pc}{PC}{Pacini}
\newacronym{pdk}{PDK}{Process Development Kit}
\newacronym{pd}{PD}{Phase Detector}
\newacronym{pll}{PLL}{Phase-Locked Loop}
\newacronym{posfet}{POS-FET}{Piezoelectric Oxide Semiconductor Field Effect Transistor}
\newacronym{prs}{PRS}{Production Rule Set}
\newacronym{pr}{P\&R}{Place and Route}
\newacronym{psc}{PSC}{Post Synaptic Current}
\newacronym{ptat}{PTAT}{Proportional To Absolute Temperature}
\newacronym{pv}{PV}{Parietal Ventral Area}
\newacronym{qdi}{QDI}{Quasi Delay Insensitive}
\newacronym{qfp}{QFP}{Quad-Flat Package}
\newacronym{ra1}{RA}{Rapid-Adapting I}
\newacronym{ra2}{RA2}{Rapid-Adapting II}
\newacronym{rafr}{RaFr}{Radio Frequency}
\newacronym{ram}{RAM}{Random Access Memory}
\newacronym{rbssg}{RBSSG}{Reverse Bitwise Synthetic Spike Generator}
\newacronym{rf}{RF}{Receptive Field}
\newacronym{rl}{RL}{Reinforcement Learning}
\newacronym{rnn}{RNN}{Recurrent Neural Network}
\newacronym{roi}{ROI}{Region Of Interest}
\newacronym{rpe}{RPE}{Reward Prediction Error}
\newacronym{s1}{S1}{Somatosensory Primary Cortex}
\newacronym{s2}{S2}{Somatosensory Secondary Cortex}
\newacronym{sa1}{SA1}{Slow-Adapting I}
\newacronym{sa2}{SA2}{Slow-Adapting II}
\newacronym{sdsp}{SDSP}{Spike-Driven Synaptic Plasticity}
\newacronym{sfd}{SFD}{Spike Frequency Divider}
\newacronym{shf}{S-HF}{Spike-based Hold \& Fire}
\newacronym{sig}{S-IG}{Spike-based Integrate \& Fire}
\newacronym{simd}{SIMD}{Single Instruction, Multiple Data}
\newacronym{slpf}{S-LPF}{Spike-based Low Pass Filter}
\newacronym{smd}{SMD}{Surface Mount Device}
\newacronym{smu}{SMU}{Source Measurement Unit}
\newacronym{soc}{SoC}{System on Chip}
\newacronym{sota}{SOTA}{State-Of-The-Art}
\newacronym{so}{SO}{Second Order}
\newacronym{spice}{SPICE}{Simulation Program with Integrated Circuit Emphasis}
\newacronym{spi}{SPI}{Serial Peripheral Interface}
\newacronym{spll}{sPLL}{Spiking Phase-Locked Loop}
\newacronym{src}{SRC}{Sparse Representation Classifier}
\newacronym{srdp}{SRDP}{Spike-Rate-Dependent Plasticity}
\newacronym{srm}{SRM}{Simple Response Model}
\newacronym{stdp}{STDP}{Spike Timing Dependent Plasticity}
\newacronym{stp}{STP}{Short Term Plasticity}
\newacronym{svm}{SVM}{Support Vector Machine}
\newacronym{syn}{SYN}{Synaptic Trace}
\newacronym{tc}{TC}{Temporal Contrast}
\newacronym{tde}{TDE}{Time Difference Encoder}
\newacronym{tpu}{TPU}{Tensor Processing Unit}
\newacronym{trg}{TRG}{Trigger Trace}
\newacronym{ttfs}{TTFS}{Time to first spike}
\newacronym{uc}{$\mu$C}{Microcontroller}
\newacronym{vco}{VCO}{Voltage Controlled Oscillator}
\newacronym{vpd}{VP-d}{Victor-Purpura Distance}
\newacronym{wta}{WTA}{Winner Take All}

\newacronym{alif}{ALIF}{Adaptive \myglsentry{lif}}
\newacronym{ann}{ANN}{Artificial \myglsentry{nn}}
\newacronym{asic}{ASIC}{Application Specific \myglsentry{ic}}
\newacronym{cmos}{CMOS}{Complementary \myglsentry{mos}}
\newacronym{cnn}{CNN}{Convolutional \myglsentry{nn}}
\newacronym{cstdp}{C-STDP}{Calcium \myglsentry{stdp}}
\newacronym{ddpi}{DDPI}{Double \myglsentry{dpi}}
\newacronym{dnn}{DNN}{Deep \myglsentry{nn}}
\newacronym{eai}{Edge-AI}{Edge \myglsentry{ai}}
\newacronym{enn}{ENN}{event-based \myglsentry{nn}}
\newacronym{epsc}{EPSC}{Excitatory \myglsentry{psc}}
\newacronym{exlif}{ExLIF}{Exponential \myglsentry{lif}}
\newacronym{fefet}{FeFET}{Ferroelectric \myglsentry{fet}}
\newacronym{gpgpu}{GPGPU}{General-Purpose computing on \myglsentry{gpu}}
\newacronym{ipsc}{IPSC}{Inhibitory \myglsentry{psc}}
\newacronym{moscap}{MOSCAP}{\myglsentry{mos} Capacitor}
\newacronym{mosfet}{MOSFET}{\myglsentry{mos} \myglsentry{fet}}
\newacronym{nmnist}{N-MNIST}{Neuromorphic \myglsentry{mnist}}
\newacronym{nmos}{nMOS}{n-type \myglsentry{mos}}
\newacronym{oxram}{OxRAM}{Oxide-based resistive \myglsentry{ram}}
\newacronym{pka}{PKA}{\myglsentry{camp}-depended Protein Kinase}
\newacronym{pmos}{pMOS}{p-type \myglsentry{mos}}
\newacronym{rram}{RRAM}{Resistive \myglsentry{ram}}
\newacronym{sadc}{sADC}{spiking \myglsentry{adc}}
\newacronym{scnn}{sCNN}{spiking \myglsentry{cnn}}
\newacronym{snn}{SNN}{Spiking \myglsentry{nn}}
\newacronym{sodpi}{SoDPI}{Second-order \myglsentry{dpi}}
\newacronym{sram}{SRAM}{Static \myglsentry{ram}}
\newacronym{tstdp}{T-STDP}{Triplet \myglsentry{stdp}}

\newacronym{adexlif}{AdExLIF}{Adaptive Exponential Leaky Integrate-and-Fire}
\newacronym{nfet}{n-FET}{\myglsentry{nmos} \myglsentry{fet}}
\newacronym{pfet}{p-FET}{\myglsentry{pmos} \myglsentry{fet}}

\newacronym{sma}{SMA}{SubMiniature version A}
\newacronym{wrota}{WR-OTA}{Wide Range \gls{ota}}
\newacronym{sstdp}{S-STDP}{Stochastic \myglsentry{stdp}}
\newacronym{neuop}{NeuOp}{Neuron spike Operation}
\newacronym{synop}{SynOp}{Synaptic Operation}
\newacronym{fdsoi}{FDSOI}{Fully Depleted Silicon On Insulator}
\newacronym{qif}{QIF}{Quadratic \myglsentry{if}}
\newacronym{ff}{FF}{edge triggered Flip-Flop}
\newacronym{lrs}{LRS}{Low Resistance State}
\newacronym{hrs}{HRS}{High Resistance State}

\makeglossaries

\begin{document}
\maketitle
\let\thefootnote\relax\footnotetext{\textsuperscript{1}Bio-Inspired Circuits and Systems (BICS) Lab, Zernike Institute for Advanced Materials, University of Groningen, Netherlands.\\
\textsuperscript{2}Groningen Cognitive Systems and Materials Center (CogniGron), University of Groningen, Netherlands.\\
\textsuperscript{3}Micro- and Nanoelectronic Systems (MNES), Technische Universit\"at Ilmenau, Germany.\\
\textsuperscript{4}Forschungszentrum Jülich, Germany.\\
\textsuperscript{5}Institute
of Neuroinformatics, University of Zürich and ETH Zürich, Switzerland.\\
\textsuperscript{6}D-ITET Integrated Systems Laboratory, ETH Zürich, Zürich, Switzerland.\\
\textsuperscript{*}Corresponding authors \{h.r.greatorex, e.chicca\}@rug.nl}

\begin{abstract}

Recent advances in memory technologies, devices and materials have shown great potential for integration into neuromorphic electronic systems. 
However, a significant gap remains between the development of these materials and the realization of large-scale, fully functional systems. 
One key challenge is determining which devices and materials are best suited for specific functions and how they can be paired with CMOS circuitry. 
To address this, we introduce TEXEL, a mixed-signal neuromorphic architecture designed to explore the integration of on-chip learning circuits and novel two- and three-terminal devices. 
TEXEL serves as an accessible platform to bridge the gap between CMOS-based neuromorphic computation and the latest advancements in emerging devices. 
In this paper, we demonstrate the readiness of TEXEL for device integration through comprehensive chip measurements and simulations.
TEXEL provides a practical system for testing bio-inspired learning algorithms alongside emerging devices, establishing a tangible link between brain-inspired computation and cutting-edge device research.

\end{abstract}


\twocolumn

\section{Introduction}

\begin{figure*}[hbt]
    \centering
    \includegraphics[width=\linewidth]{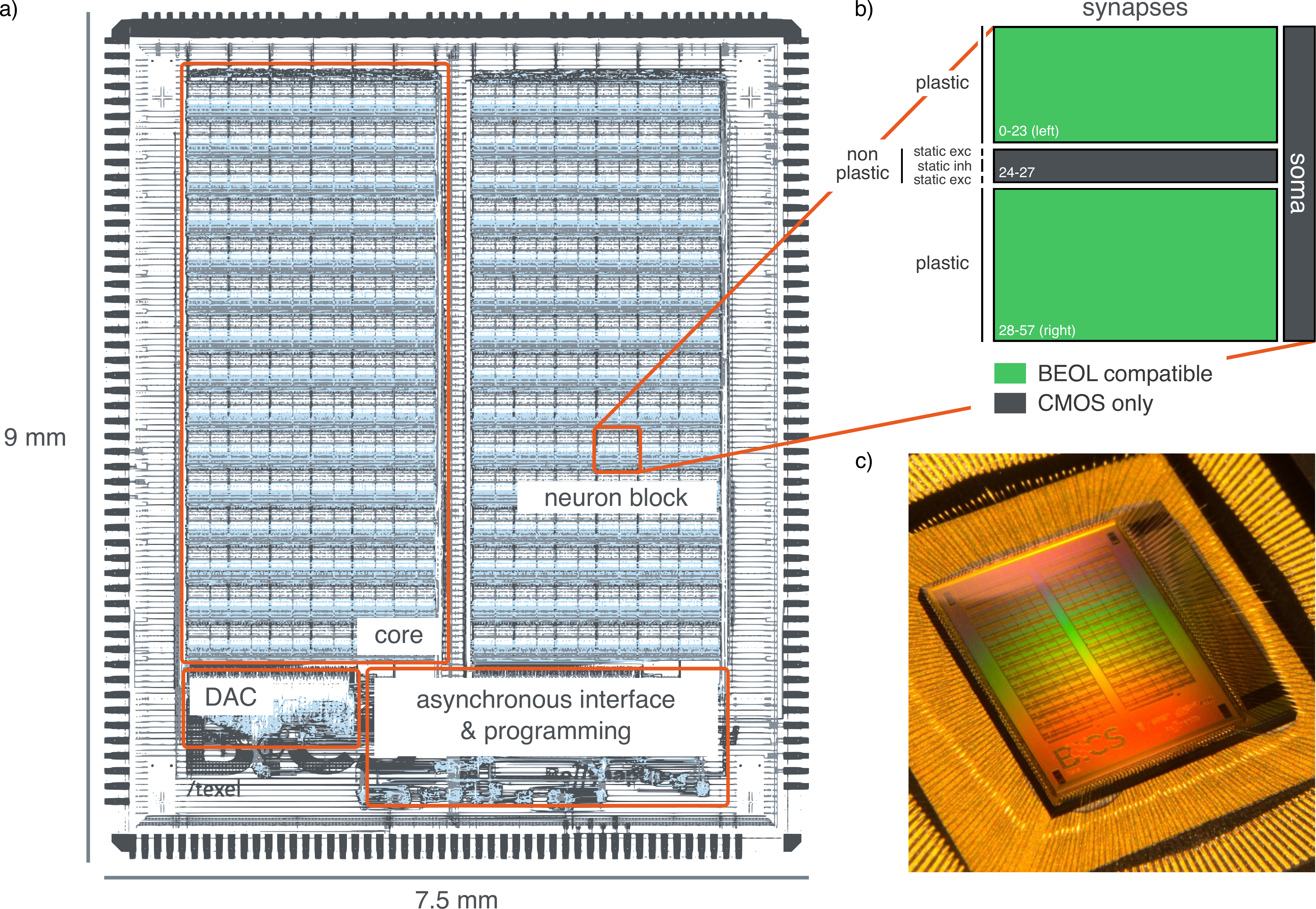}
    \vspace{10pt}
    \caption{\textbf{The fabricated TEXEL chip.} \textbf{a)} Footprint of the chip, indicating the location of the architectural blocks. 
    \textbf{b)} The neuron block footprint, indicating the synaptic fan-in of the soma within the block. 
    The location of the plastic and non-plastic synapses are shown with excitatory (exc) and inhibitory (inh) types. 
    The plastic synapses contain the contacts and interface circuitry for BEOL integration of memristive devices. 
    \textbf{c)} A photograph of the \qty{9}{\mm} $\times$ \qty{7,5}{\mm} die, fabricated using the XFAB~\qty{180}{\nano\meter} process.}
    \label{fig:overview}
\end{figure*}

The unsustainable energy requirements of current deep learning algorithms have promoted research into novel computing architectures and technologies. 
Some of these efforts are aimed at emulating the computational principles of biological intelligence to enhance efficiency and processing capabilities. 
In this regard, the development of neuromorphic computing architectures has seen substantial growth~\cite{richter_dynap-se2_2024, pehle2022brainscales, gonzalez_spinnaker2_2024, orchard_efficient_2021, Davies_etal18, Moradi_etal18, benjamin_neurogrid_2014}. 
In particular, neuromorphic systems using hybrid CMOS-memristive circuits offer a promising direction for low-power, highly compact \gls{imc} solutions~\cite{sebastian2020memory,christensen20222022}.
Memristive technologies encompass a wide range of novel electronic materials and devices that possess inherent memory and reprogrammability through state-dependent, and possibly non-volatile, resistance modulation~\cite{yang2013memristive}.

When integrated in CMOS \gls{snn} chips, hybrid neuromorphic/memristive circuits can exploit the physics of the devices and their intrinsic dynamics to carry out low-power computations that extend the basic advantages of conventional \gls{imc} dense crossbar array architectures~\cite{Chicca_Indiveri20_doi}.
Conventional \gls{imc} neural network designs, which use either digital \gls{sram} or memristive crossbars, aim to maximize peak throughput and area efficiency~\cite{hongyang2021programmable,keller2022accelerator,khaddam2022hermes}.
In contrast, mixed-signal neuromorphic architectures seek to reduce overall power consumption, especially in edge computing applications like bio-signal processing or environmental monitoring, which involve slowly varying signals~\cite{donati_etal19}. 
Recent research has focused on brain-inspired neural mechanisms to implement efficient neural networks targeting edge-computing applications~\cite{fra_2022,gallou_24}.
These types of architectures implement \glspl{snn}, where the spikes are digital events communicated via asynchronous digital logic. 
Both the analog circuits implementing the slow neural and synaptic dynamics as well as the asynchronous digital circuits implementing the event-based routing and network (re)programmability, enable ultra-low power computation. 
Typically, the analog circuits used in these neuromorphic platforms rely on the subthreshold analog transistor regime~\cite{Liu_etal02b_doi} to emulate neuron-like dynamics for a further reduction in energy cost~\cite{Chicca_etal14b, Moradi_etal18, rubino2021}.

By exploiting the physics of the devices, this approach has led to the development of a diverse array of circuits that implement computational models of synaptic plasticity~\cite{khacef2023spike_doi}.
Synaptic plasticity is the ability of synapses to be potentiated or depressed in a volatile (short-term plasticity) or non-volatile manner (long-term plasticity)~\cite{george2020}.
Although pure CMOS hardware implementations of local synaptic plasticity rules have been shown to express complex and powerful computational properties~\cite{Qiao_etal15,Chicca_etal03,pehle2022brainscales}, they require substantial silicon real-estate to store the synaptic weights.
Addressing this issue has traditionally involved a common strategy: driving the weight to a stable value for storage. 
The use of bistable plastic synapses originates from some of the first developments of full-scale neuromorphic systems~\cite{Indiveri_etal06, Chicca_etal03} mimicking biological synapses which inherently have limited bit precision~\cite{Bartol_2025,Amit_Fusi94a_doi}. 
Other works propose to update the weights directly within a digital memory~\cite{cartiglia_22, pehle2022brainscales}, thus facilitating a long-term storage, but they often require a continuous power supply to maintain the memory.
Combining the mixed-signal neuromorphic engineering approach with the integration of memristive devices, would simultaneously enable the exploration of additional computational strategies, such as intrinsic stochasticity and state-dependence, as well as provide a compact and non-volatile storage option for maintaining weight values during power-cycles.

Recent efforts have thus initiated the exploration of integrating memristive devices with CMOS neuromorphic systems, aiming to leverage the synergy of both technologies~\cite{dagostino2023denram, zhang_hybrid_2021, Payvand_2020_b, mesa_cmol2022, Nair_etal17, zhao2020memristor, wang_fully_2018}.
A majority of these efforts have focused on complementing memristive crossbar arrays with neuromorphic peripheral circuitry to handle the generation of output spikes and the computation of learning signals~\cite{Payvand_2020_b, zhang_hybrid_2021, mesa_cmol2022}.
Synaptic weights in these systems are realized by the resistance states of memristive devices in a crossbar array. 
To modify these weights, suitable read and write processes must be developed, which can compromise the systems' ability to perform \gls{imc}.
On the other hand, few works explore the possibility of implementing \gls{imc} synaptic plasticity, with learning directly occurring at each synaptic device.
In~\cite{Nair_etal17} the authors proposed a differential three-terminal device interface to achieve more flexible device access for online learning while~\cite{zhao2020memristor} and~\cite{wang_fully_2018} proposed the exploitation of memristive device dynamics to implement in-memory plasticity directly in the crossbar. 
Although these approaches have been explored, they have been limited to simulations of a few circuit elements with restricted learning flexibility.

In this work, we introduce TEXEL, a fully fabricated chip combining the operational efficiencies of memristive devices with the spike-based approach of neuromorphic systems. 
The chip exploits the analog subthreshold CMOS regime and event-based computation to implement ultra-low power spiking neurons and plastic synapses with tunable always-on trace-based local learning functionality.
TEXEL incorporates a novel \gls{beol} device-agnostic differential synaptic interface, enabling the integration of a wide range of two- and three-terminal memristive devices across all ~9K plastic synapses atop the CMOS chip. 
This design makes TEXEL a versatile research platform for large-scale \gls{beol} device integration in neuromorphic systems.
While devices are yet to be integrated, TEXEL represents, to the best of our knowledge, the first memristor-based large-scale neuromorphic chip with on-chip learning that is fully-fabricated. 
It exploits the synergies of \gls{imc} and spiking neural networks to present a concrete step towards the following key developments for such systems:

 \begin{enumerate}
\item Exploiting the capability of memristive materials and devices to facilitate the implementation and consolidation of on-chip synaptic plasticity.
\item Providing a platform to explore the large-scale \gls{beol} integration of memristive materials and devices with an \gls{snn} processor. 
\end{enumerate}

Here, we present the architectural innovations and learning mechanisms of the TEXEL chip, highlighting its impact on the ongoing development of neuromorphic computing and the pursuit of beyond-CMOS device integration.
Through comparisons with other existing full-scale neuromorphic chips, we highlight TEXEL's unique contributions and envisage its role in advancing the frontier of computing toward more efficient, brain-inspired paradigms.

\section{Results}

Silicon measurements that validate the functionality of the TEXEL chip \mbox{(Fig.~\ref{fig:overview})} and are outlined in the following sections.
Experiments using the on-chip learning circuits demonstrate the emergent phenomena of \gls{stdp}~\cite{Song_etal00} and \gls{srdp}~\cite{Brader_etal07}.
The functionality of memristive device read-write circuits are verified experimentally, and the operation of the interfacing circuits is demonstrated with post-layout simulations, which define the parameter range of memristive devices aiming for compatibility with TEXEL.
Power consumption measurements provide a detailed breakdown of the contribution of each circuit block, exemplifying the inherent power efficiency advantages of subthreshold analog circuitry and the event-driven paradigm.

\subsection{Neural Circuits}
\begin{figure*}
    \centering
    \begin{subfigure}[b]{0.3\textwidth}
        \centering
        \includegraphics[width=\textwidth]{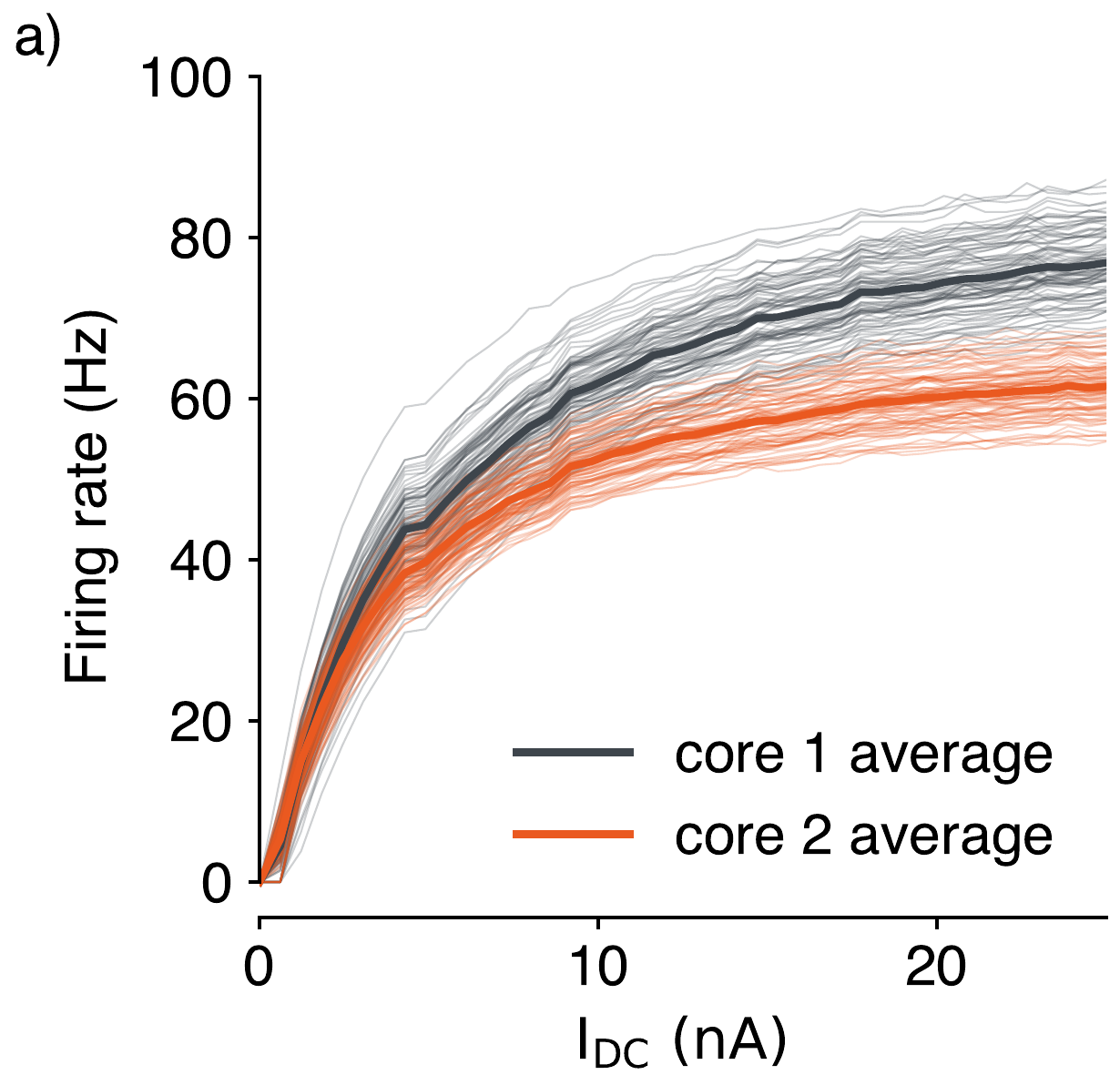}
    \end{subfigure}
    \hfill
    \begin{subfigure}[b]{0.3\textwidth}
        \centering
        \includegraphics[width=\textwidth]{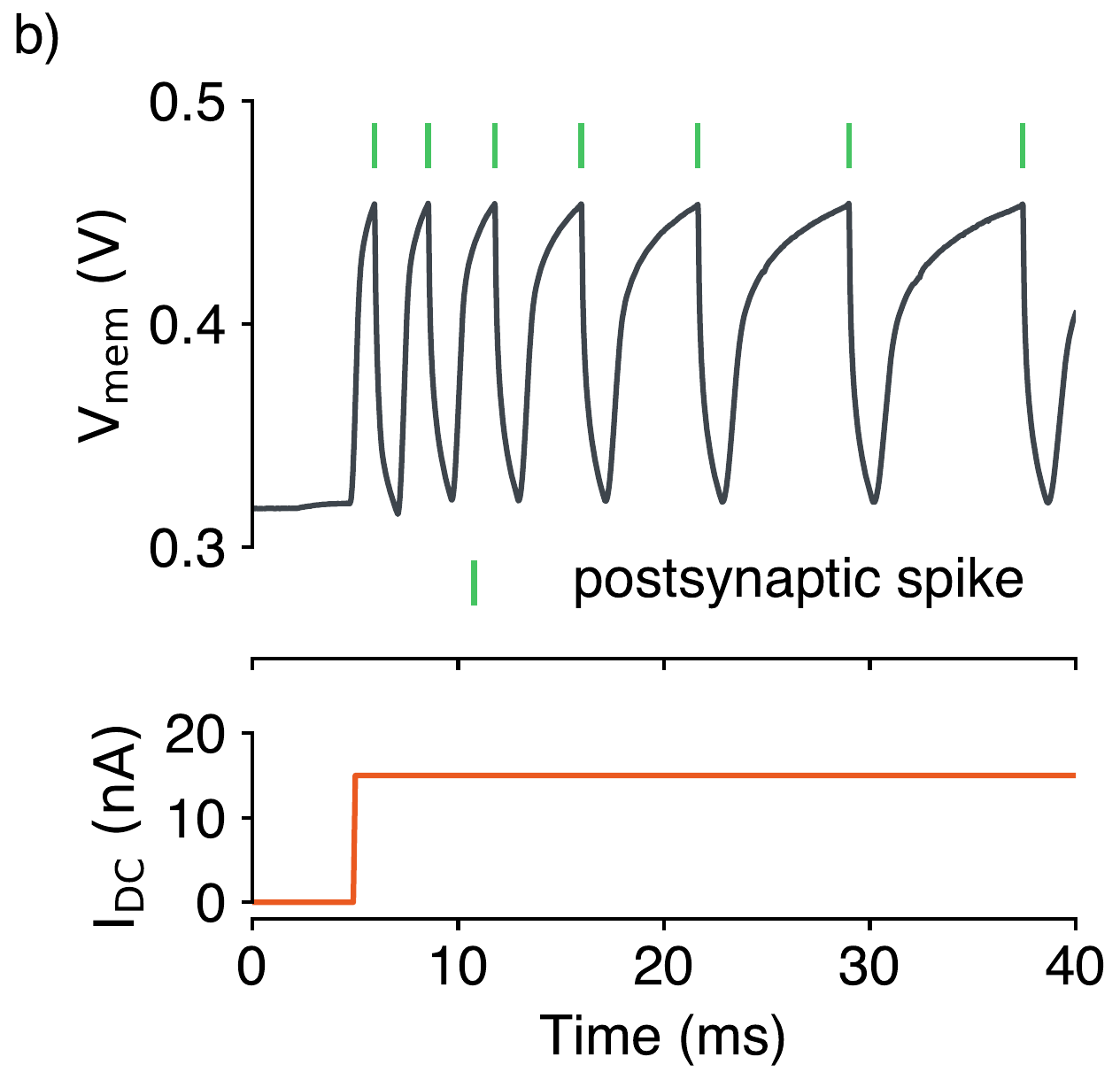}
    \end{subfigure}
    \hfill
    \begin{subfigure}[b]{0.3\textwidth}
        \centering
        \includegraphics[width=\textwidth]{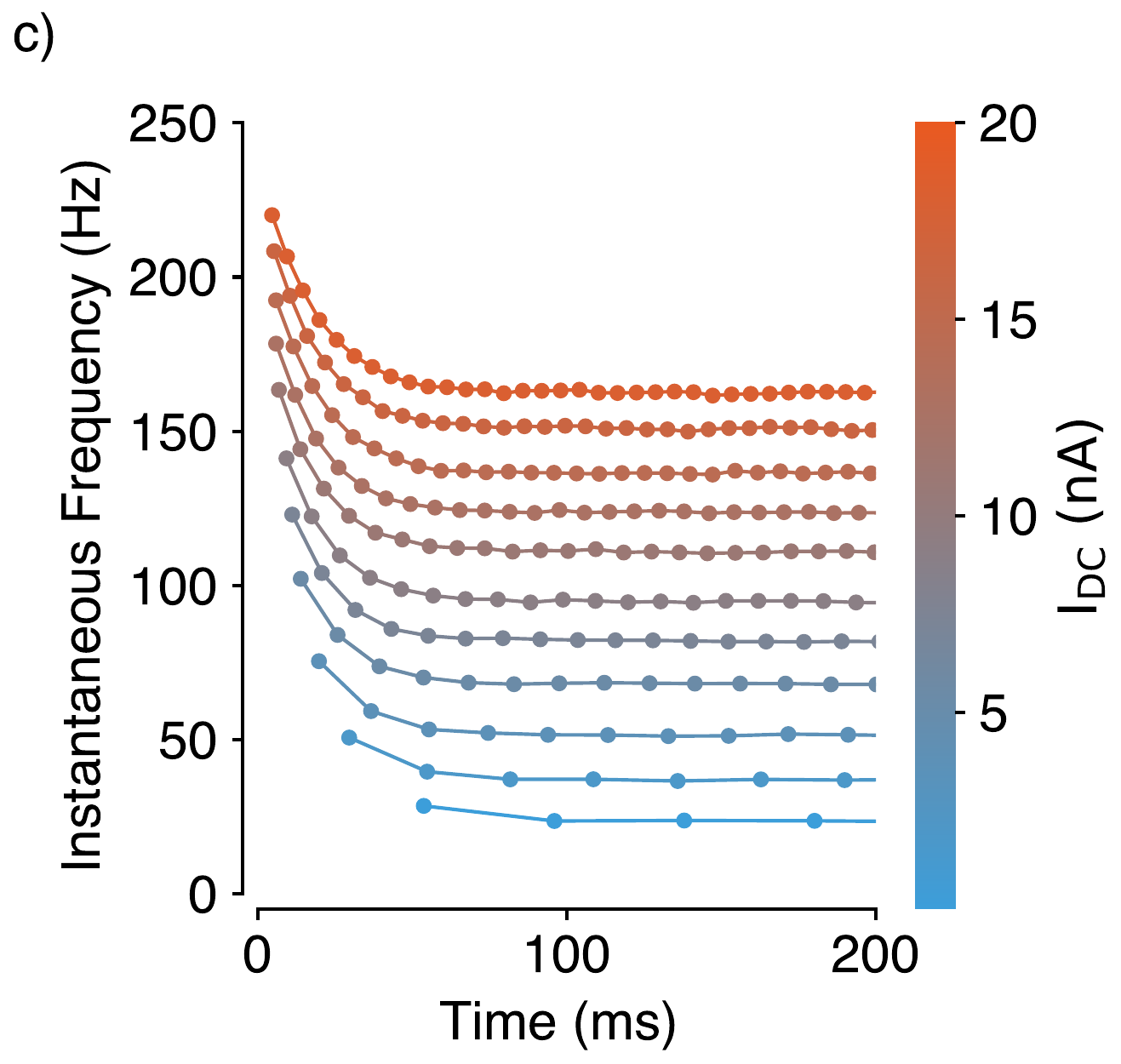}
    \end{subfigure}
    \begin{subfigure}[b]{\textwidth}
    \centering
    \vspace{10pt}
    \includegraphics[width=\linewidth]{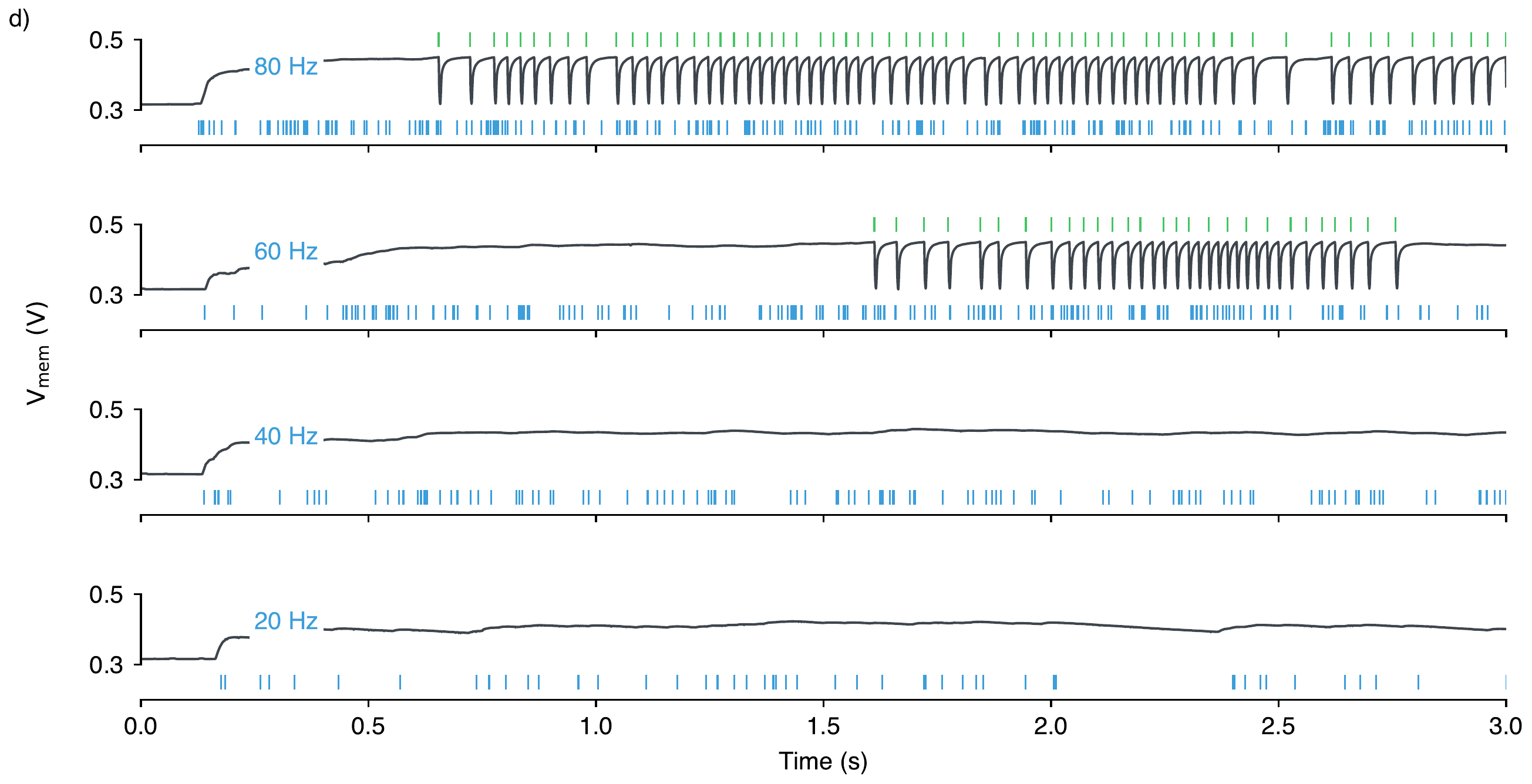}
    \end{subfigure}
    \caption{\textbf{Measurements of the neuron circuitry on the TEXEL chip.} 
    \textbf{a)} The measured firing rates of all neurons on TEXEL in response to a constant \gls{dc} input current.
    \textbf{b)} Measurement of the membrane potential of a single neuron in response to a \gls{dc} step input. 
    The neuron's adaptation characteristic is evident as its firing rate begins high and gradually diminishes to attain a steady state. 
    \textbf{c)} Measurements of the variations in the instantaneous firing rate and timing of output spikes in relation to the magnitude of \gls{dc} injected into the soma. 
    \textbf{d)} The recorded membrane potential response of a neuron receiving presynaptic Poisson input at the static excitatory synapses. 
    Below, in blue, is the presynaptic spike train, while above, in green, the postsynaptic spikes indicate the neuron's spiking activity.}
    \label{fig:neuron_measurements}
\end{figure*}

We measured the activation of the silicon neuron circuits to assess their transfer function and operating regimes.
A \gls{dc} input was applied and systematically increased across all neurons while their spike rate was recorded (\mbox{Fig.~\ref{fig:neuron_measurements}a}).
The resultant \gls{fi} curve shows both the aggregate mean response for each core as well as the individual activation profiles of all neurons. 
The discernible core-specific disparity is attributable to mismatch in the biasing circuitry.
The dispersion in the \gls{fi} curve of each neuron stems from inherent variations in individual neuron circuits.
While device mismatch variability can be reduced by including calibration procedures for each element~\cite{pehle2022brainscales}, we chose to minimize it, through judicious analog circuit design techniques, and keep it, as it can be exploited for example in learning~\cite{perez2021neural}.

We conducted validation measurements of the adaptive characteristics embedded in the circuitry of each neuron. 
The response of the membrane potential to a \gls{dc} step input was measured, as well as the timing of output spikes (\mbox{Fig.~\ref{fig:neuron_measurements}b}).
These observations reveal the expected temporal pattern in the neuron's instantaneous spike rate, characterized by an initial peak followed by a gradual decay towards a stable state (\mbox{Fig.~\ref{fig:neuron_measurements}c}). 
Figure~\ref{fig:neuron_measurements}d shows an instance in which a neuron is stimulated by a Poisson spike train through its static excitatory synapses.
The plastic synapses located within each neuron block are quantized to a binary value which is translated into an analog bias representing high and low synaptic efficacy. 
The on-chip weight matrix, encoding the state of all plastic synapses, can be read-out post learning and also programmed for inference (see Supplementary \mbox{Fig.~\ref{fig:inference}}).

\subsection{Learning Circuits}

The on-chip plasticity was implemented using mixed-signal circuitry embedded within each plastic synapse.
This circuitry emulates the \gls{bcall} rule~\cite{willian_etal2024}, which combines \gls{stdp} for low activity with Hebbian changes~\cite{Hebb49} at high activity. 
In this model, synaptic updates are driven by pre- and postsynaptic calcium traces representing neuronal activity.
A secondary postsynaptic trace ($\mathrm{Ca}^{2+}$) with a slower time constant acts as a plasticity gating mechanism, ensuring weight updates occur only within specific firing rate ranges.
The learning rule imposes a bistable analog internal weight ($V_w$) to help mitigate catastrophic forgetting in binary synapses~\cite{Senn_Fusi05, Mitra_etal09, Brader_etal07, rubino23}, stabilizing synaptic states using accumulated updates and bistability circuitry. 

\begin{figure*}
    \centering
    \begin{subfigure}{\textwidth}
        \centering
        \includegraphics[width=\textwidth]{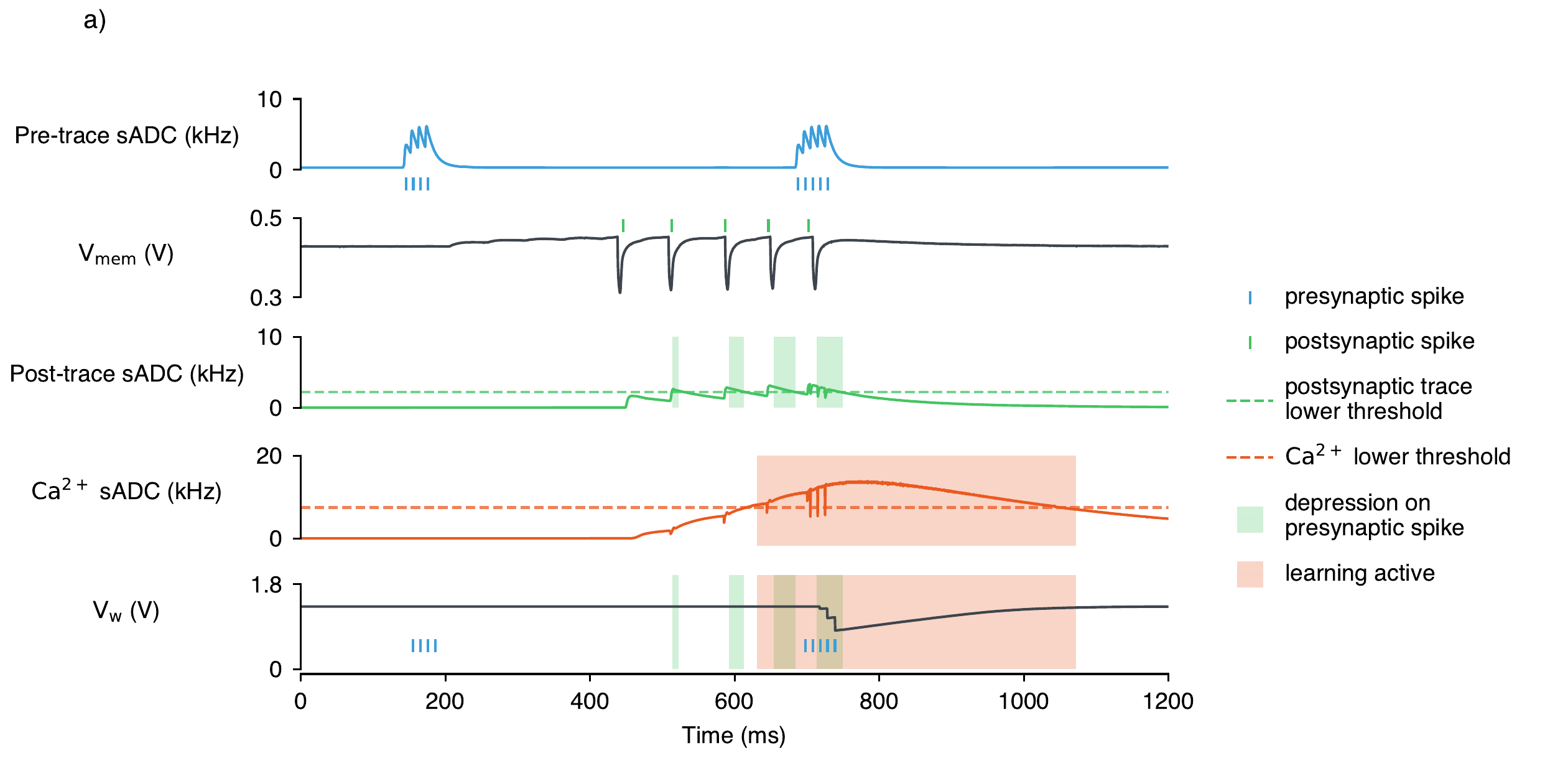}
    \end{subfigure}
    \hfill
    \begin{subfigure}[b]{\textwidth}
        \centering
        \includegraphics[width=\textwidth]{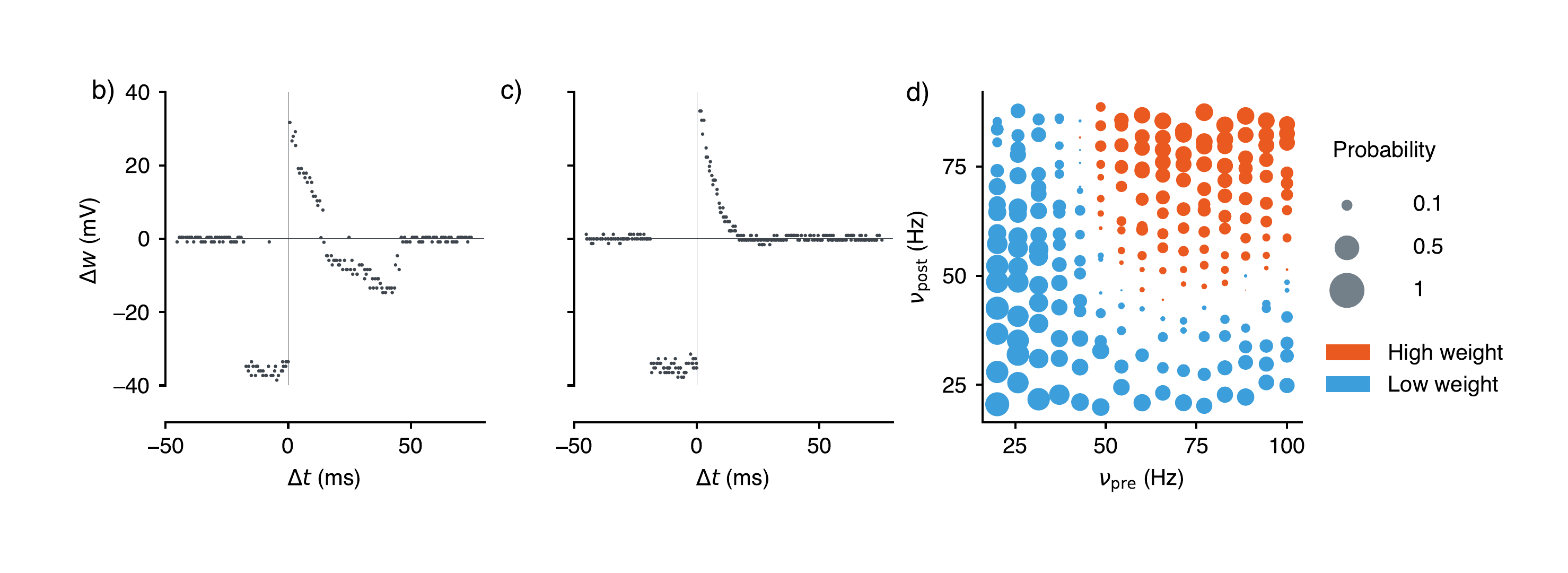}
    \end{subfigure}
    \vspace{-20pt}
    \caption{
    \textbf{Silicon measurements of a single plastic synapse and its neuron, demonstrating local synaptic plasticity (complementary to Fig.~\ref{fig:texel_soma_synapse_circuits}b).}
    \textbf{a)} A presynaptic spike train induces a current (blue) read by the \gls{sadc}, while simultaneous stimulation with an \gls{epsc} triggers postsynaptic spiking (green). 
    Shaded regions indicate when the post-trace exceeds the lower threshold, reflecting short-term memory. 
    The $\mathrm{Ca}^{2+}$ trace (orange) accumulates postsynaptic activity, showing plasticity when above its threshold.
    \textbf{b)} \gls{stdp} measurements assess the impact pre- and postsynaptic spike timing, $\Delta t = \mathrm{pre} - \mathrm{post}$, on the analog weight of the synapse ($w$).
    \textbf{c)} This \gls{stdp} curve demonstrates modulation of potentiation and depression through analog biasing.
    \textbf{d)} \gls{srdp} results show the probability of the synapse having high or low weight based on pre- and postsynaptic firing rates ($\nu_{\text{pre}}$ and $\nu_{\text{post}}$).
    }
    \label{fig:learning_results}
\end{figure*}

\mbox{Figure~\ref{fig:learning_results}a} shows measurements of a plastic synapse undergoing short-term depression. 
Pre-trace integration of presynaptic spikes maintains a decaying record of presynaptic activity, but without postsynaptic activity, the synaptic weight remains unchanged. 
When postsynaptic spikes occur, plasticity becomes apparent, and if the post-trace crosses it's lower threshold, depression is triggered. 
The synaptic weight experiences short-term depression but stabilizes to the high state due to bistability circuitry.

We conducted in-silico experiments to characterize \gls{stdp} of the learning circuitry (Figs.~\ref{fig:learning_results}b,~\ref{fig:learning_results}c). 
The synaptic weight changes ($\Delta w$) were measured by systematically varying pre- and postsynaptic spike timings.
Adjusting the biasing parameters allowed for on-chip configuration of \gls{stdp} curves, enabling the introduction of depressive regions for positive pre-post pairings.
Additionally, \gls{srdp} was measured by varying pre- and postsynaptic Poisson spike rates. 
A probability map (Fig.~\ref{fig:learning_results}d) of synaptic weight changes demonstrates that under conditions of high presynaptic and postsynaptic activity, the likelihood of the synapse settling into a potentiated (high) state increases. 
In contrast, when activity levels are lower, the synapse is more likely to undergo depression, favoring the low-weight state. 
This data highlights the sensitivity of the learning circuitry to the frequency and timing of local spiking activity.

\subsection{Memristive Device Interfacing Circuits}

\begin{figure*}
    \centering
    \includegraphics[width=\textwidth]{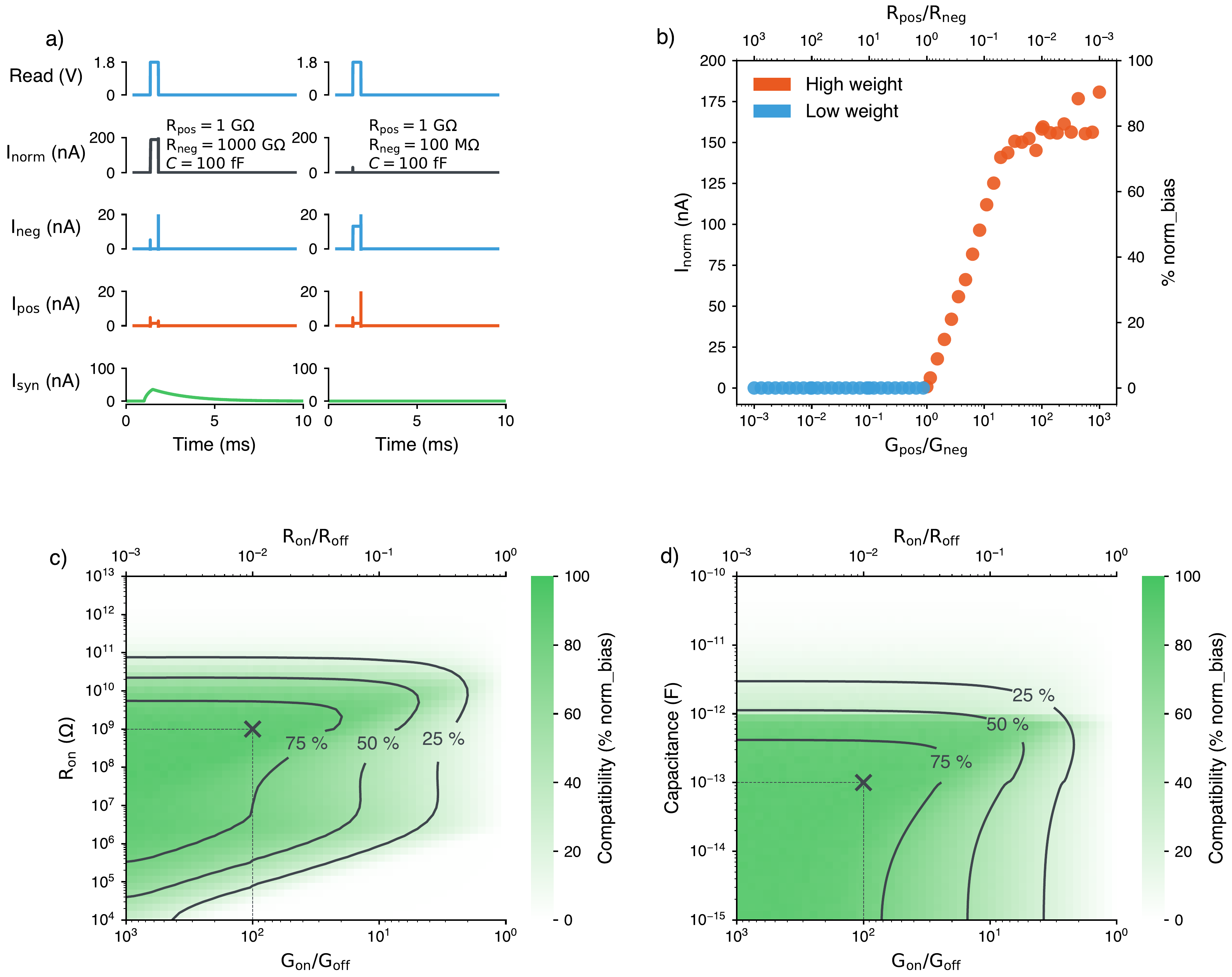}
    \vspace{5pt}
    \caption{
    \textbf{Spectre post-layout simulations of the read protocol for the differential normalizer synapse on TEXEL.}
    \textbf{a)} A read pulse with a width of \qty{500}{\micro s} activates the normalizer circuit, sourcing $I_{\mathrm{neg}}$ and $I_{\mathrm{pos}}$. 
    The circuit outputs a non-zero current, $I_{\mathrm{norm}}$, if $I_{\mathrm{pos}} > I_{\mathrm{neg}}$, which is integrated by a \gls{dpi} synapse, resulting in a current $I_{\mathrm{syn}}$ sent to the neuron. 
    The left panel shows high weight storage ($\mathrm{R}_{\mathrm{pos}} < \mathrm{R}_{\mathrm{neg}}$), eliciting a response, while the right panel shows low weight storage ($\mathrm{R}_{\mathrm{neg}} < \mathrm{R}_{\mathrm{pos}}$), where no current is integrated.
    \textbf{b)} With $R_{\text{pos}}=\qty{1}{\giga \ohm}$, device capacitance of $C=\qty{100}{\fF}$, and a read pulse width of \qty{500}{\us}, the relative resistances of both devices are varied by sweeping $R_{\text{neg}}$. 
    The average output current of the normalizer circuit is measured as a \% of $\mathrm{norm\_bias}$, showing non-zero current when the positive device's conductance exceeds that of the negative device.
    \textbf{c)} Simulations explore device characteristics' impact on compatibility with TEXEL. 
    The cross ($\times$) represents a device with $C = \qty{100}{\fF}$, $\mathrm{G}_\mathrm{on} / \mathrm{G}_\mathrm{off}=100$, $\mathrm{R}_{\text{on}} = \qty{1}{\giga \ohm}$, and a read pulse width of \qty{500}{\us}. 
    Heatmaps indicate average current from the normalizer as a percentage of $\mathrm{norm\_bias}$.
    \textbf{d)} A sweep of the device's capacitance versus its on/off ratio is shown with $\mathrm{R}_{\text{on}}$ fixed at \qty{1}{\giga \ohm}.
    }
    \label{fig:device_compatibility}
\end{figure*}

Each plastic synapse on TEXEL \mbox{(Fig.~\ref{fig:overview}b)} can be enabled to utilize a pair of memristive devices to store a binary weight using a differential device configuration~\cite{Bocquet_etal18, Hirtzlin_etal_20}.
When the chip is programmed to enable device operation, at the time of a presynaptic spike, the synaptic weight is read using a differential normalizer circuit~\cite{Nair_etal17}.
To demonstrate the operation of the normalizer circuitry we performed extensive Spectre post-layout simulations over a range of memristive device parameters, namely: conductance, capacitance and on-off ratio.
The memristive devices were modelled as parallel RC circuits. 
\mbox{Figure~\ref{fig:device_compatibility}a} shows how the differential device setup, consisting of a ``positive'' and ``negative'' device, is able to store the binary synaptic weight. 
In the case where the resistance of the positive device is lower than that of the negative device, the current sourced through the positive device, $I_\mathrm{pos}$, during a read pulse (presynaptic spike) is greater than the current sourced through the negative device, $I_\mathrm{neg}$.
In this scenario the normalizer circuit transmits a current, $I_\mathrm{norm}$, proportional to the biasing of the normalizer circuit, $\mathrm{norm\_bias}$. 
For these simulations the current was normalised to \qty{200}{\nano\ampere} ($\mathrm{norm\_bias}$) and passed into a \gls{dpi} synapse~\cite{Bartolozzi_Indiveri07b} to elicit a postsynaptic current, $I_\mathrm{syn}$.
In the alternative case, when the differential synapse is programmed to represent a low weight, the positive device resistance is greater than the negative device resistance.
Therefore $I_\mathrm{neg} > I_\mathrm{pos}$ and the normalizer circuit does not convey a current. 
\mbox{Figure~\ref{fig:device_compatibility}b} presents post-layout simulation results showing how $I_\mathrm{norm}$ varies with the ratio of the positive and negative device conductance. 
When the ratio is $< 1$, $I_\mathrm{norm}$ is zero, conversely when the ratio is $> 1$, $I_\mathrm{norm}$ is large enough to elicit a postsynaptic current. 
For large ratios between the positive and negative devices, translating to a large on-off ratio, the differential synapse and normalizer circuit is able to source a current that is closer to $\mathrm{norm\_bias}$.

\subsection{Memristive Device Requirements}

To quantify the compatibility of TEXEL with co-integrated memristive devices we performed extensive Spectre post-layout simulations of the CMOS interface circuitry with realistic device characteristics, over several orders of magnitude. 
We parameterised all simulations using a fixed read voltage pulse width of \qty{500}{\us} and a $\mathrm{norm\_bias}$ of \qty{200}{\nA}, however these can be varied using the on-chip programming and biasing. 
\mbox{Figure~\ref{fig:device_compatibility}c} shows a heat-map of a 2D logarithmic device characteristic sweep during which the on-off conductivity ratio of the device was varied with the on-resistance. 
This heat-map shows the percentage of the $\mathrm{norm\_bias}$ of the normalizer circuit that was sourced during a read voltage pulse that was sent to the differential device synapse when storing a high weight. 
This is used as the metric determining whether a memristive device will operate as expected when integrated with the TEXEL chip and defines the ``compatibility''.
Similarly, we performed simulations varying the on-off conductivity ratio and capacitance of the device \mbox{(Fig.~\ref{fig:device_compatibility}d)}, here the same metric of compatibility is used. 
This is an additional memristive device constraint that must be satisfied to ensure successful integration with CMOS and one that is often overlooked. Table~\ref{tab:device_table} presents the integration specifications derived from the aforementioned simulations, operating voltages and circuit footprints. 

\begin{table*}
\centering
\begin{tabular}{ccccc} 
\toprule
     &\textbf{Read Voltage} &\textbf{Set Voltage} &\textbf{Reset Voltage} & \textbf{Area} \\ 
 & (\qty{}{\volt}) & (\qty{}{\volt}) & (\qty{}{\volt}) & (\qty{}{\micro\meter\squared}) \\
\midrule
\rule{0pt}{10pt}
Min. & 0               & -5                          & -5               &  -             \\ 
\rule{0pt}{10pt}
Max. & 5               & 5                           & 5                &        114         \\ 
\bottomrule \\
     &                  &                              &                   &                 \\
\toprule
     & \textbf{Total Capacitance}  & \textbf{Capacitance/Area} & $\mathbf{R_{\text{on}}}$ & $\mathbf{G_{\text{on}} / G_{\text{off}}}$    \\
 & (\qty{}{\pico\farad}) &( \qty{}{{\farad\per\centi\meter\squared}}) & (\qty{}{\giga\ohm}) & - \\
\midrule
\rule{0pt}{10pt}
Min. & -                & -                            & -                 & 10              \\ 
\rule{0pt}{10pt}
Max. & 10            & \num{8.8e-7}                  & 10             & -               \\
\bottomrule
\end{tabular}
\vspace{10pt}
\caption{\textbf{TEXEL memristive device compatibility requirements for integration.} Entries in the second row are derived from post-layout simulations (Fig.~\ref{fig:device_compatibility}), 50\% is taken as a confidence threshold for compatibility.}
\label{tab:device_table}
\end{table*}

\subsection{Power Measurements}
\begin{figure*}
    \centering
    \includegraphics[width=\textwidth]{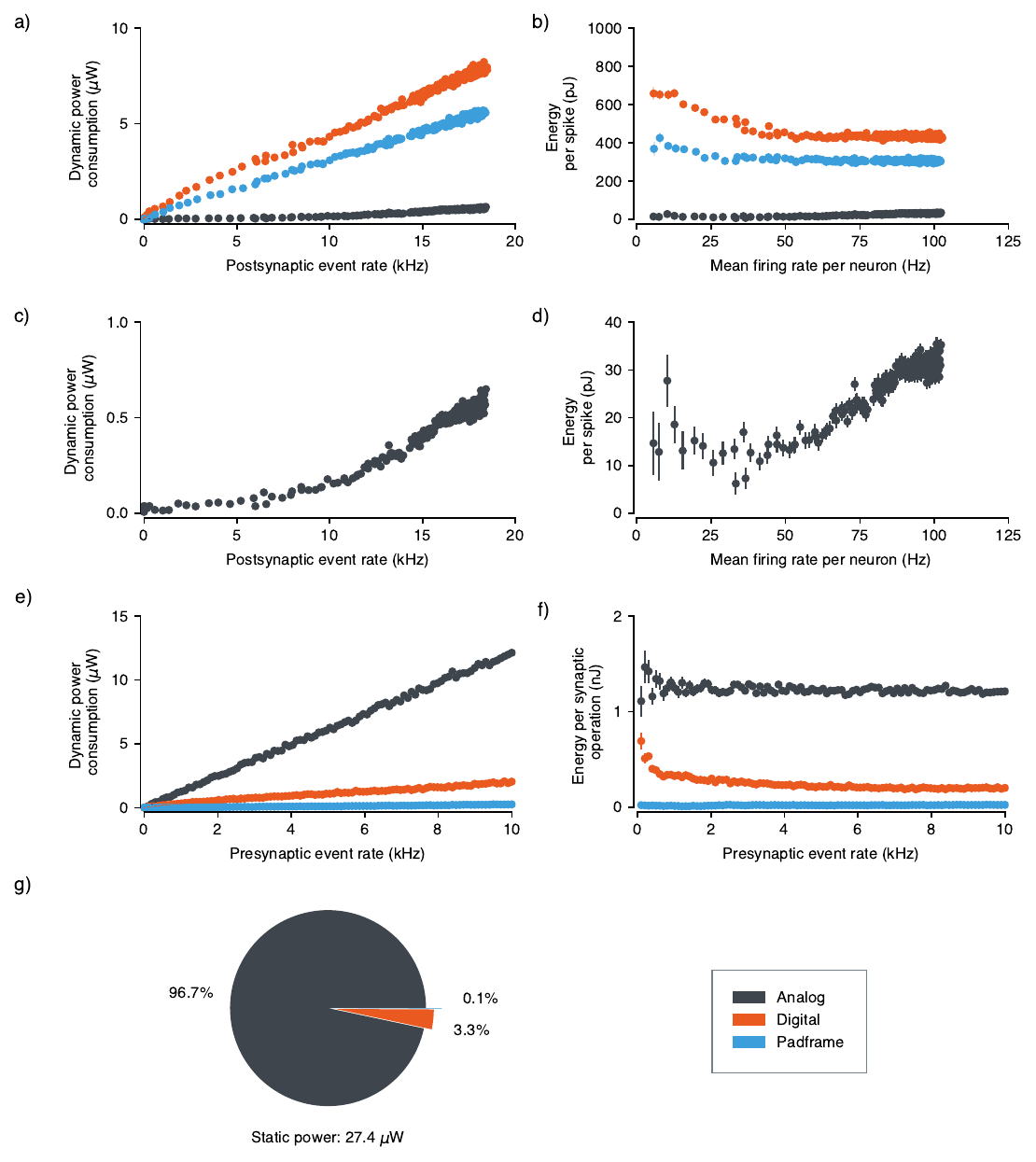}
    \caption{
    \textbf{Dynamic and static power measurements of the TEXEL chip, focusing on energy consumption for synaptic operations and neuron spikes.}
    \textbf{a)} Dynamic power consumption versus postsynaptic event rate, measured for the three isolated power supplies.
    \textbf{b)} Energy per spike for increasing mean firing rates across each power supply.
    \textbf{c)} Dynamic power consumption of the analog power supply against postsynaptic event rate.
    \textbf{d)} Energy consumed per spike versus mean firing rate per neuron, for the analog power supply.
    \textbf{e)} Dynamic power consumption during random synaptic stimulation at increasing input event rates.
    \textbf{f)} Energy consumption per synaptic operation against input event rate.
    \textbf{g)} Breakdown of static power consumption while neurons are inactive and synapses are unstimulated.
    }
    \label{fig:power_measurements}
\end{figure*}

We conducted extensive power measurements on the TEXEL chip using a femtoampere \gls{smu} to assess its power distribution across operations for the analog and digital power sources. 
\mbox{Figure~\ref{fig:power_measurements}a and \ref{fig:power_measurements}c} show how the dynamic power consumption varies with the global spike rate of the chip, this was modulated by increasing the \gls{dc} input bias for all neurons. 
The total dynamic power consumption is divided into the contributions of the isolated digital, analog and padframe power supplies. 
The energy per spike was also calculated for varying spiking rates \mbox{(Fig~\ref{fig:power_measurements}b, Fig~\ref{fig:power_measurements}d)}. \mbox{Figure~\ref{fig:power_measurements}e and \ref{fig:power_measurements}f} show the same power contribution breakdown for synaptic operations with the energy required per operation. 
This experiment was performed by increasing the input spike rate over the \gls{aer} bus, randomly addressing all synapses on the chip, over both cores. 
\mbox{Figure~\ref{fig:power_measurements}g} shows the breakdown of the static power consumption of the chip, measured at \qty{27,4}{\uW}.

\section{Discussion}

Recent advancements have produced only a few successfully co-integrated large-scale memristor-CMOS neuromorphic systems~\cite{yan_rram-based_2019,wan_331_2020,wan_compute--memory_2022,de_los_rios_multi_core_2023,valentian2019fully}, with most relying on foundry assistance~\cite{wan_331_2020,wan_compute--memory_2022,de_los_rios_multi_core_2023,valentian2019fully}.
This lack of co-integration is a key challenge in advancing memristor-CMOS systems, emphasizing the importance of wafer-level integration platforms to bridge the gap with CMOS technology and to progress neuromorphic chip development.

In this work, we introduced TEXEL, an \gls{snn} processor with on-chip learning circuits capable of interfacing with a large range of memristive device operation requirements (Table.~\ref{tab:device_table}). 
TEXEL functions in both full-CMOS and device-integrated modes, offering a versatile platform to explore emerging memristive technologies within the context of a spiking neuromorphic system. 
The platform supports a wide range of device interfacing options, including read-write pulse widths from \qty{10}{\ns} to \qty{100}{\ms}, continuous read and pre-charge modes (see Supplementary Section~\ref{appendix:dev_operation}), and high-voltage compatibility up to \qty{5}{\volt}. 
It can interface with either two-terminal devices or three-terminal devices such as \mbox{\glspl{fefet}~\cite{falcone2022back,dutta2022logic}}.

The chip also provides a crucial substrate for testing yield and endurance - key factors in device fabrication at scale.
When incorporated into \gls{snn} systems, these properties can be evaluated in real-world learning tasks, while on-chip learning algorithms enable continuous performance characterization through repeated weight updates. 
This iterative process provides valuable insights into the ability of algorithms to mitigate drift in device characteristics and maintain performance over time.
While TEXEL’s broad compatibility with \gls{nvm} devices provides flexibility, it also introduces significant area overhead (see Table~\ref{tab:texel_comp_table} and Supplementary Fig.~\ref{fig:neuron_block}). 
Future iterations, once a specific \gls{nvm} technology is chosen, should aim to optimize both density and performance. 
TEXEL prioritizes flexibility over efficiency by supporting multiple device debug modes and allowing operation without devices for CMOS circuit verification. 
Nonetheless, the ability to monitor a wide range of signals remains essential for benchmarking and debugging hybrid memristor-CMOS systems (see Supplementary Table~\ref{tab:monitoring}).

Broadly, the integration of memristive devices and materials with CMOS extends beyond storing and reading synaptic weights; they can also be incorporated into neuron~\cite{choi2020} and learning circuits~\cite{demirag21}, enhancing characteristics such as time constants. 
Moreover, these emerging devices and materials have shown significant promise in sensory applications~\cite{dai2023, lenk2023}, rendering them particularly appealing for integration into sensory front ends that can be interfaced with always-on neuromorphic chips. 
They have also been shown to facilitate the implementation of additional network features, such as synaptic delays~\cite{dagostino2023denram} and specific network topologies~\cite{dalgaty2024}, further enhancing the richness and versatility of neuromorphic systems. 
Collectively, these capabilities position memristive device technology as a key component in the development of efficient and adaptable electronic architectures.
With its flexibility, TEXEL serves as a foundational tool for expediting the realization of memristor-CMOS systems, paving the way for scalable, state-of-the-art spiking neural network chips that can effectively leverage emerging device technologies.

\begin{table*}
\centering
\begin{tabular}{l*{4}{c}}
\toprule
 \textbf{Chip} & \textbf{TEXEL} & \textbf{ISSCC'20} & \textbf{ISCAS'23} & \textbf{NeuRRAM}\\
     &[this work]& \cite{wan_331_2020} & \cite{de_los_rios_multi_core_2023} & \cite{wan_compute--memory_2022}\\
     \midrule

    Design & mixed-signal & mixed-signal & mixed-signal &  mixed-signal\\ 
    CMOS technology & \qty{180}{\nano\meter} & \qty{130}{\nano\meter} & \qty{130}{\nano\meter} & \qty{130}{\nano\meter}\\
    Device type & \textbf{any \acrshort{beol} current based} & \acrshort{rram} & \acrshort{oxram} &  \acrshort{rram} \\
    Device terminals & \textbf{2-3} & 2 & 2 & 2\\
    Number of devices & 19\,k & 65\,k & 4\,k & 3.14\,M\\
    Area including I/O & \qty{67.5}{\milli\meter\squared} & - & - & \qty{158.76}{\milli\meter\squared} \\
    Core area & \qty{44.98}{\milli\meter\squared} & \qty{1.79}{\milli\meter\squared} & \qty{0.180}{\milli\meter\squared} & -\\
    \midrule
    Neuron model &  \acrshort{adexlif} & \acrshort{if} & \acrshort{if} &  \acrshort{if}\\
    Number of neurons & 180 & 256 & 64 & 12\,k \\
    Number of synapses & 10\,k & 65\,k & 4\,k & 3.14\,M\\
    Full parallel write & yes & - & column-wise & -\\
    In-memory plasticity$^*$ & \cmark & \xmark & \xmark & \xmark \\
    Learning rule & \textbf{\acrshort{stdp} \& \acrshort{sdsp}} & - & \acrshort{sstdp} & -\\
    \midrule
    Energy/spike \acrshort{neuop} &  \qty{25.9}{\pico\joule} @ \qty{80}{\hertz} & \qty[per-mode=symbol]{0.0139}{\pico\joule\per\mac} & - & \qty{0.121}{\pico\joule} \\
    \bottomrule
\\
\toprule
 \textbf{Chip} & \textbf{NElec'18} & \textbf{IEDM'19} & \textbf{VLSIT'19}\\
     & \cite{wang_fully_2018} & \cite{valentian2019fully} & \cite{yan_rram-based_2019} \\
     \midrule
    Design  & memristor & mixed-signal & mixed-signal\\ 
    CMOS technology & - & 130\,nm & 150\,nm\\
    Device type & HfO$_x$ \acrshort{rram} & \acrshort{oxram}& HfO$_x$ \acrshort{rram}\\
    Device terminals & 2 &2 &2\\
    Number of devices & 74 & 13.5\,k & 64\,k\\
    Area including I/O & \qty{0.56}{\milli\meter\squared} & - & -\\
    Core area & - & - & - \\
    \midrule
    Neuron model  & stochastic \acrshort{lif}  & \acrshort{if} & \acrshort{if} \\
    Number of neurons & 8 & 10 & 256\\
    Number of synapses & 64 & 1440 & 65\,k\\
    Full parallel write & yes & - & -\\
    In-memory plasticity & \cmark & \xmark & \xmark \\
    Learning rule & Hebbian LTP & - & -\\
    \midrule
    Energy per spike/\acrshort{neuop} & - & - & \qty[per-mode=symbol]{0.257}{\pico\joule\per\mac}\\
    \bottomrule
\end{tabular}
\begin{tablenotes}
    \item * The plasticity rule is implemented in-memory with local circuits, instead of off-crossbar generation of learning signals, eliminating off-array communication.
\end{tablenotes}
\caption{\textbf{Comparison of TEXEL with other silicon-verified memristor-\gls{snn} chips.}}
\label{tab:texel_comp_table}
\end{table*}

\section{Methods}

\subsection{Chip Architecture}

With 2 cores of 90 neuron blocks, TEXEL hosts 180 neurons each with 58 complex synapses \mbox{(Fig.~\ref{fig:overview}b)}. 
The chip's digital periphery operates asynchronously, utilizing handshake protocols between functional blocks~\cite{ataei_2021}. 
Robustness was tested through extensive testing for variable switching delays, eliminating the reliance on specific timing constraints.
Spike I/O and register operations share an asynchronous pipeline tailored for \gls{aer}. 
Demux circuits route incoming packets to either the spike decoder or register block.
The decoder translates external \gls{aer} spike packets, while the encoder processes on-chip neuron spikes for transmission off-chip. 
The register block comprises 64 23-bit asynchronous memory arrays (per core) used for biasing and programming, each capable of parallel read or write operations. 
All analog circuitry is biased using a 12-bit \gls{dac} (see Supplementary Section~\ref{appendix:dac}). 
To enable the integration of two- and three-terminal \gls{nvm} devices there is interfacing circuitry including terminal contacts placed within each plastic synapse in every neuron block~\cite{neckar_braindrop_2019, park_22-pjspike_2023, richter_dynap-se2_2024} (see Supplementary Fig.~\ref{fig:neuron_block}). 
Figure~\ref{fig:overview} shows the embedding of the neuron blocks and synapses within the chip architecture.

\subsection{Neuron Circuits}

\begin{figure*}
    \centering
    \includegraphics[width=\textwidth]{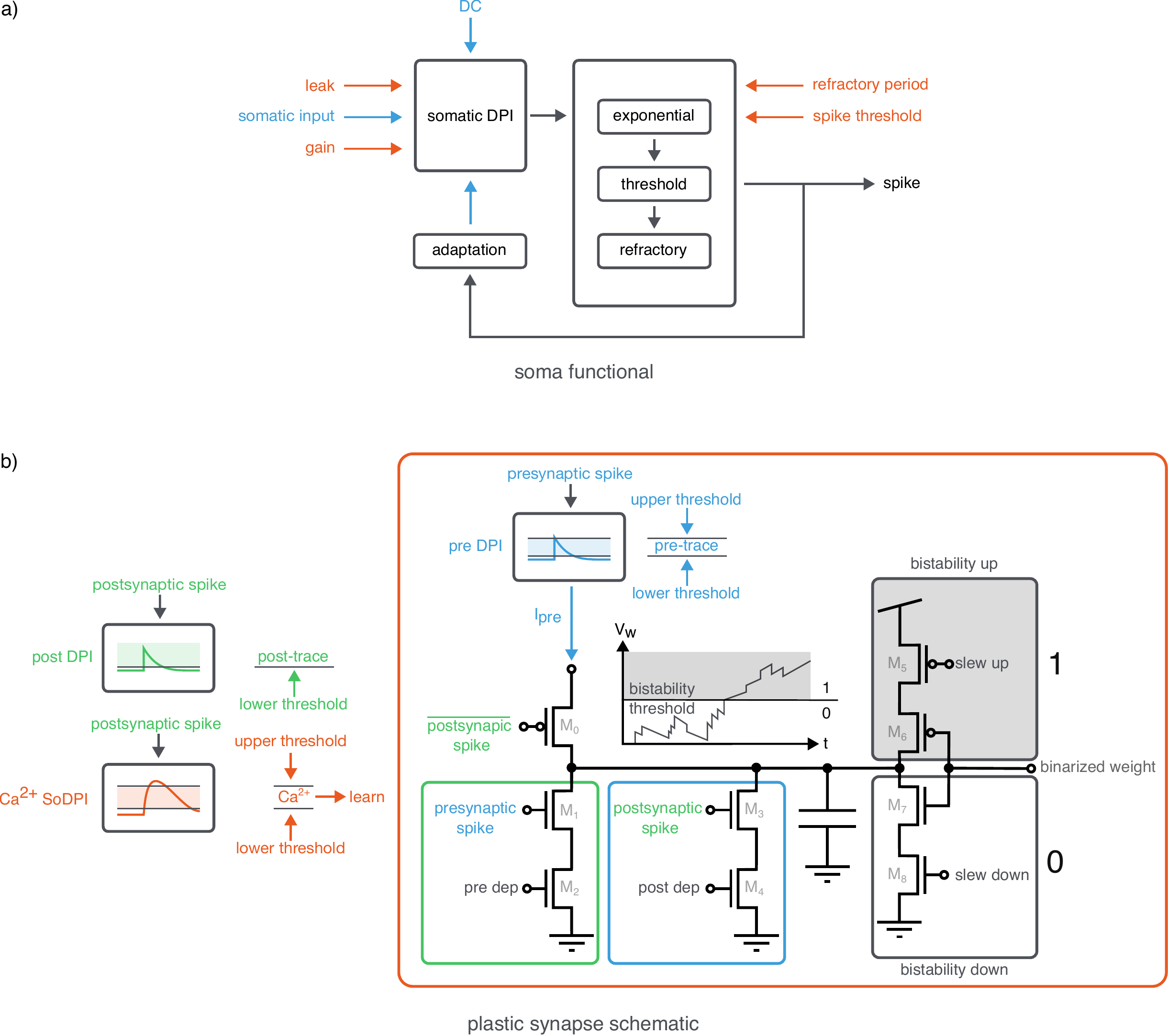}
    \vspace{5pt}
    \caption{
    \textbf{The functional architecture of the neuron on the TEXEL chip and schematic of the plasticity circuit in each plastic synapse.}
    \textbf{a)} The neuron features a somatic \gls{dpi} that integrates input from both DC and synaptic sources, with circuits for thresholding, spike generation, refractory period, and positive feedback to mimic biological spiking neurons. 
    An adaptation mechanism can be enabled to modulate spike frequency. 
    Orange inputs represent tunable biases, and blue elements indicate current sources.
    \textbf{b)} The plasticity circuit in each synapse uses three analog traces to govern weight updates. 
    Two neuron-level traces, the postsynaptic trace (post-trace) and the $\mathrm{Ca}^{2+}$ trace, are transmitted to synapses and must meet threshold conditions for weight updates. 
    If the post-trace exceeds a threshold, incoming presynaptic spikes reduce the synaptic weight by a fixed increment. 
    The presynaptic activity (pre-trace) also determines whether the weight will increase or decrease, with updates occurring via charge deposition on a capacitor. 
    The weight is then quantized into high or low states by a bistability circuit, which controls drift toward ground or supply voltage, with drift rates set by the $\mathrm{slew \ up}$ and $\mathrm{slew \ down}$ biases.
    }
    \label{fig:texel_soma_synapse_circuits}
\end{figure*}

The \gls{adexlif} neuron circuit integrated on TEXEL is the latest iteration of a continuing design evolution that has undergone multiple enhancements to optimize performance~\cite{Moradi_etal18, Indiveri03, Livi_Indiveri09_updated, Chicca_etal14b, Qiao_etal16, richter_dynap-se2_2024}. 
The implementation of the neuron draws inspiration from the improvements detailed in~\cite{rubino2021}, focusing on minimizing power consumption and reducing mismatch.
The neuron dynamics are driven by two inputs: a DC input and a somatic input current from the synaptic fan-in, enabling network-level experiments.
\mbox{Figure~\ref{fig:texel_soma_synapse_circuits}a} details the \gls{adexlif} circuit, showing its distinct functional blocks. 
A somatic input \gls{dpi} models the neuron’s leak conductance, integrating synaptic currents into the membrane capacitance, producing a membrane current representing the neuron state variable~\cite{Bartolozzi07}. 
Between the somatic \gls{dpi} and spike generation, three modules control membrane current dynamics: a threshold, exponential and refractory module. 
The threshold module, implemented with a low-power current comparator, triggers a spike at the moment the membrane current exceeds the spiking threshold. 
The exponential module, implemented with a current-based positive feedback, accelerates the membrane current increase when it is closer to the spiking threshold.
Once the neuron generates a spike, the refractory module keeps the neuron silent for a certain time set by the refractory period bias.
Furthermore, there is an adaptation module, implemented with a pulse extender and a negative feedback low-pass filter circuit (\gls{dpi}). 
This is activated with each output spike event, integrating the neuron's recent spiking activity.
All aforementioned modules can be controlled using seven tunable biases.
The neuron circuit is designed to be compatible with \gls{aer} circuits therefore an asynchronous digital handshaking block is incorporated to transmit spikes as address-events through the \gls{aer} pipeline.

\subsection{Synaptic Circuits}
Each neuron on the TEXEL chip has a synaptic fan-in of 58 synapses, 54 plastic and 4 non-plastic (static). 
Non-plastic synapses are realised through \gls{dpi} circuits and activate in response to a presynaptic spike, producing a current with an amplitude that is tunable. 
Consequently, they can be deactivated by setting the weight bias current to zero. 
The nature of the non-plastic synapses is predetermined, with two per-neuron designated as excitatory and two as inhibitory (\mbox{Fig.~\ref{fig:overview}b}). 
The weight of the plastic synapses, updated according to the on-chip local learning rule, is stored on a capacitor on a short-time scale and discretized into two stable states on a long-time scale. 
The weight update occurs in the analog domain, while the long-term storage takes place in the digital domain. 
The nature of the plastic synapses (excitatory or inhibitory) can be configured on-chip.
Excitatory synapses inject a positive current into the soma, while inhibitory ones draw current away from it.
The total synaptic activity, computed as the sum of weighted currents, is transmitted to four different \glspl{dpi}, each independently tunable. 

\subsection{Learning Circuits}
Within each plastic synapse there exists a CMOS implementation of the \gls{bcall} rule~\cite{Willian_etal2024_hardware} that can be enabled, making use of signals local to each synapse to facilitate either Hebbian or anti-Hebbian \gls{sdsp} (Fig.~\ref{fig:texel_soma_synapse_circuits}b). 
A pre-trace, realised by a \gls{dpi} circuit~\cite{Bartolozzi_Indiveri07b}, maintains a decaying memory of the presynaptic spike train. 
If this trace exists between an upper and lower threshold then with the cooccurance of postsynaptic spikes the synaptic weight is depressed. 
In parallel, depression can also occur if the post-trace, a short term memory of postsynaptic activity, is above a low threshold and a presynaptic spike occurs. 
Potentiation occurs on a postsynaptic spike during which the value of the presynaptic trace is sampled from such that the magnitude of potentiation is proportional to the presynaptic trace at that time~\cite{Indiveri_etal06}. 
A smooth third trace, realised by a \gls{sodpi} circuit~\cite{richter_subthreshold_2023}, is used to track the neurons' activity, representing the postsynaptic neuron's $\mathrm{Ca}^{2+}$ concentration.
The upper and lower thresholds of the $\mathrm{Ca}^{2+}$ trace establish a "stop-learning" region, restricting synaptic plasticity to occur only within this range. 
The weight is stored as a voltage as shown in Figure~\ref{fig:learning_results}a and is discretized via a voltage threshold. 
Additionally, a bistability circuit is employed such that over the long time scale the weight drifts towards a binary value. 
The temporal dynamics of the aforementioned traces, strength of the potentiation/depression events, bistability slew rates and thresholds can all be varied through the biasing of the analog circuitry. 

\begin{figure*}
    \centering
    \includegraphics[width=\linewidth]{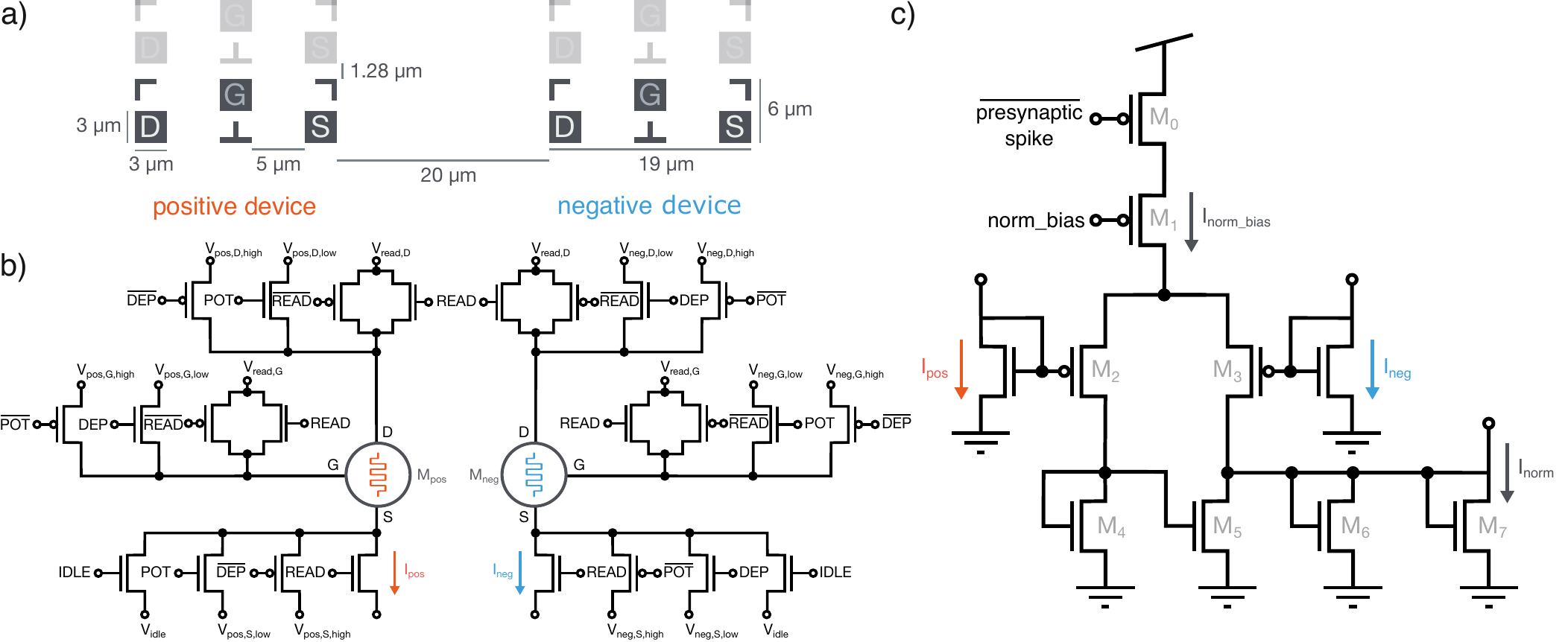}
    \vspace{5pt}
    \caption{\textbf{The footprint and schematic of the per-synapse device interface terminals and schematic of the differential normalizer circuit.} 
    \textbf{a)} A diagram illustrating the physical dimensions and spatial arrangement of the source, drain, and gate contacts for two- or three-terminal devices. 
    Each synapse deploys two devices configured differentially, serving as both positive and negative components. 
    The diagram also provides information on the spacing between synaptic rows, depicting the distances between adjacent devices in each synapse. 
    \textbf{b)} Schematic of device interface circuitry. 
    All voltages can be set in the range \qtyrange{0}{5}{\volt} in order to read or write both devices in the differential configuration. 
    \textbf{c)} The differential normalizer circuit functions to compare the currents generated by positive and negative devices during a device read, prompted by a presynaptic spike. 
    It evaluates the disparity between these currents and generates an output current, denoted as $I_\mathrm{norm}$, which is proportional to the normalized discrepancy between $I_\mathrm{pos}$ and $I_\mathrm{neg}$. 
    Moreover, $I_\mathrm{norm}$ is exclusively non-zero when $I_\mathrm{pos}$ surpasses $I_\mathrm{neg}$ and can be modulated by the bias $\mathrm{norm\_bias}$. 
    Consequently, the output represents the binary state of the synapse, and the sourced current is directed towards a \gls{dpi} circuit for further processing.}
    \label{fig:device_interface}
\end{figure*}

\subsection{Memristive Device Integration}

To support large-scale integration of plastic memristor-based synapses, the chip is designed with a ``device-agnostic'' architecture, ensuring high flexibility and offering multiple probing configurations for different memristive devices. 
This design accommodates both two- and three-terminal devices, supporting a broad range of operating voltages and currents \mbox{(see Table~\ref{tab:device_table})}. 
Device behavior can be monitored either through on-chip read-outs of output currents during operation or via off-chip access to all device terminals through the interface circuit (see Supplementary Table~\ref{tab:monitoring}). 
Full access to the device terminals enables external burn-in or programming of the memristive devices.



To facilitate \gls{beol} integration, each terminal is accessible through a high-level metal contact with spacing and sizing depicted in \mbox{Figure~\ref{fig:device_interface}a}.
Three branches in the interface circuitry employ n- and p-type transistors, along with transmission gates, to deliver voltage pulses for reading device states or for potentiating or depressing synaptic weights \mbox{(Fig.~\ref{fig:device_interface}b)}. 
The operation voltages are provided off-chip as inputs to the padframe with a maximum voltage of \qty{5}{\volt}. 
Digital signals to the transistor gates are internally controlled by a synapse controller circuit which implements synaptic operations and weight updates.
We note that an extra idle transistor and idle signal is used to facilitate the possibility of pre-charging the device between read pulses and allow a better distinction between their \gls{hrs} and \gls{lrs} currents (see Supplementary Section~\ref{appendix:dev_operation}).

Often, device operation specifications are not immediately compatible with the technology node and cannot be compensated for by voltage scaling or pulse length modulation. 
This can occur when currents are too low or too high, device variability is significant, or the resulting output ranges are undefined. 
In these cases, scaling and normalising circuits can be employed. 
Given this initial assumption about the properties of a device aiming for compatibility, the TEXEL chip uses the difference in state of two devices to store the synaptic weight of each plastic synapse.
Therefore the canonical on-chip operation protocol for memristive devices is binary and complementary.
As a result, while using the on-chip plasticity, devices are only switched in a binary operation between \gls{hrs} and \gls{lrs}, and always in a complementary fashion where if one is in its \gls{hrs}, the other will be in its \gls{lrs}.
A differential normalizer circuit is used to compare the responses of two devices~\cite{Nair_etal17} when the synaptic weight it being read.
When the synapse is addressed for a read, at a presynaptic spike, the currents are sourced from the devices, $I_{\text{pos}}$ and $I_{\text{neg}}$, and the normalizer circuit (Fig.~\ref{fig:device_interface}c) rescales and rectifies the detected difference to the output current range required by the \gls{dpi} synapse, $I_{\text{norm}}$. 
The rescaling factor of the output current can be modulated by the bias $\mathrm{norm\_bias}$.

Since many memristive devices use the same terminals for both reading and writing, they require exclusive control to prevent conflicts. 
In other words, when a read and write instruction occur simultaneously, a decision must be made regarding which operation to execute first. 
To manage this, each plastic synapse has a dedicated control circuit that ensures mutual exclusivity between read and write pulses (see Supplementary Fig.~\ref{fig:dev_logic}). 
Read instructions are prioritized, therefore if both commands occur concurrently, the write pulse is applied only after the read operation is completed.

\section{Author Contributions}

The following authors contributed significantly to the CMOS design of the TEXEL chip - A. R., E. C., G. I., H. G., J. C., M. C., M. F., M. M., O. R., P. K., W. S. G.

Conceptualisation - A. R., E. C., G. I., H. G., J. C., M. C., M. F., M. M., O. R., P. K., W. S. G.; Methodology - A. R. , E. C., G. I., H. G., J. C., M. C., M. F., M. M., O. R., P. K., W. S. G.; Software/Hardware - A. R., E. C., G. I., H. G., J. C., M. C., M. F., M. M., O. R., P. K., W. S. G.; Investigation - A. R. , E. C., G. I., H. G., M. M., O. R.; Writing - A. R., E. C., G. I., H. G., M. C., M. F., L. B. L., M. M., M. Z., O. R., P. K.; Visualisation - H. G., O. R., M. M., M.C., E. C., G. I.;  Supervision - E. C., G. I.

\section{Acknowledgements}

The authors would like to thank \mbox{Adrian Whatley} and \mbox{Herman Adema} for their technical support in developing PCBs, $\mu$C firmware and the TEXEL API. 
Additionally we would like to thank \mbox{Nicoletta Risi} and \mbox{Matei Zainea} for their investigations into algorithms and hardware. 
Thanks to \mbox{Ton Juny Pina} for his help soldering PCBs. 
Finally we would like to thank \mbox{Erika Covi}, \mbox{Luca Fehlings}, \mbox{Paolo Gibertini}, \mbox{Giuseppe Leo} and \mbox{Ton Juny Pina} for their feedback on the manuscript. 
This work has been supported by BeFerroSynaptic (871737), EU H2020 projects NeuTouch (813713) and MANIC (861153). Additional funding by the Deutsche Forschungsgemeinschaft (DFG, German Research Foundation): Project MemTDE Project number 441959088 as part of the DFG priority program SPP 2262 MemrisTec Project number 422738993; Project NMVAC Project number 432009531. 
The authors would like to acknowledge the financial support of the CogniGron research center and the Ubbo Emmius Funds (Univ. of Groningen). 
The authors would like to acknowledge the support and thank IC Manage, Inc for providing us with their Global Design Platform XL for design data management.

\section{Competing interests}
The authors declare no competing interests.

\printbibliography

\onecolumn

\newpage
\appendix
\renewcommand\thefigure{S\arabic{figure}}
\setcounter{figure}{0}
\setcounter{table}{0}
\setcounter{page}{1}

\section{Supplementary Materials}
\begin{refsection}

\subsection{Signal Monitoring}
\label{appendix:signal_monitoring}

TEXEL provides several methods to observe the internal state of various components, circuits and signals. 
There are three distinct monitoring methods shown in Table~\ref{tab:monitoring} under the ``Domain'' column. 
The first are analog outputs, they provide access to real time signals measurable by an oscilloscope or \glspl{adc}. 
The second monitoring method are digital output pins which expose internal digital signals. 
The third is the asynchronous \glspl{sadc} interface.

When using the \gls{sadc} bank, the current under observation is mirrored inside a circuit (described in depth in ~\cite{Voulgari_etal15}) that generates a spike rate proportional to the current magnitude. 
The spikes are propagated through an encoder interfaced with a dedicated 5 bit \gls{aer} bus. 
The \gls{sadc} bank allows the user to monitor 49682 currents, of which 24 simultaneously. 
Structures that convert currents into spikes for monitoring are popular solutions in literature with several known implementations~\cite{Corradi_Indiveri15,Voulgari_etal15,Qiao_Indiveri16,Ben_Benjamin23}. 
In many cases, the signal under observation can be selected across synapses and neurons. 
This is denoted by the ``Mux'' column in Table~\ref{tab:monitoring}.

Figure~\ref{fig:learning_results} offers an example of how the monitoring methods can be used. 
In the first row, for example, the presynaptic trace current ($I_{\text{PRE}}$) of a synapse is shown while it receives an input spike train. 
The synapse has been selected among all the available synapses but setting a register inside the chip through the input \gls{aer} bus. 
$I_{\text{PRE}}$ is recorded using the \gls{sadc} and the spiking activity of the \gls{sadc} is transmitted through the dedicated \gls{aer} bus. 
The spikes recorded by the \gls{uc} are used to reconstruct the dynamics of the signal. 
This is done by finding the \gls{isi} of the spike train and taking the reciprocal to calculate the instantaneous spike rate at the corresponding spike time. 
The resulting current proxy is visible in Figure~\ref{fig:learning_results}a, using the instantaneous firing rate. 
The same procedure is repeated for the  $I_{\text{POST}}$ and the $\mathrm{Ca^{2+}_{below}}$ current traces, depicted in the same Figure~\ref{fig:learning_results}a. 
It is noted that the \gls{sadc} spiking output is able to capture dynamics on three different time scales effectively (\qty{10}{\milli\second} for $I_{\text{PRE}}$, \qty{100}{\milli\second} for $I_{\text{POST}}$ and \qty{1}{\second} for the $\mathrm{Ca^{2+}_{below}}$ trace). 
In the second row of Figure~\ref{fig:learning_results}a, we see another example of the monitoring capability of the chip: $\mathrm{V_{mem}}$, the membrane voltage of the neuron. 
Using the same procedure explained for the synapse, a specific neuron is chosen for monitoring. 
This outputs the membrane potential of the neuron on a \gls{bnc} cable. 
For synaptic signals, $\mathrm{V_w}$ can be observed in the last row of \mbox{Figure~\ref{fig:learning_results}a}, by selecting it through a monitoring register.

\begin{table}
\centering
\begin{tblr}{
  row{19} = {c},
  cell{19}{1} = {c=6}{},
  hlines,
  hline{1-2, 19-21, 26} = {-}{0.08em},
  hline{3-18,21-25} = {0.00em},
}
\textbf{Name}              & \textbf{Description}                                     & \textbf{Type} & \textbf{Domain} & \textbf{Port} & \textbf{Mux} \\
${I_\text{DAC}}$           & Current of a single DAC (for calibration)                & Current       & Frequency             & sADC          & -             \\
$I_{\text{PRE}}$           & Pre trace of the plastic synapse                         & Current       & Frequency             & sADC          & SYN          \\
$I_{\text{SO}}$            & Second order trace of the $\mathrm{Ca}^{2+}$ SoDPI trace & Current       & Frequency             & sADC          & NRN          \\
$I_{\text{POST}}$          & Post trace of the neuron                                 & Current       & Frequency             & sADC          & NRN          \\
$I_{\text{P-LEFT}}$        & Current of the plastic left synapse                      & Current       & Frequency             & sADC          & SYN          \\
$I_{\text{P-RIGHT}}$       & Current of the plastic right synapse                     & Current       & Frequency             & sADC          & SYN          \\
$I_{\text{S-EXC}}$         & Current of the static excitatory synapse                 & Current       & Frequency             & sADC          & NRN          \\
$I_\text{AHP}$             & Adaptive current of the neuron                           & Current       & Frequency             & sADC          & NRN          \\
$I_{\text{FO}}$            & First order trace of the $\mathrm{Ca}^{2+}$ SoDPI trace  & Current       & Frequency             & sADC          & NRN          \\
$I_{\text{S-INH}}$         & Current of the static inhibitory synapse                 & Current       & Frequency             & sADC          & NRN          \\
$V_{\text{MEM}}$           & Membrane voltage of the neuron                           & Voltage       & Analog          & BNC           & NRN          \\
$V_{\text{W}}$             & Analog weight of plastic synapse                         & Voltage       & Analog          & BNC           & SYN          \\
$I_{\text{DAC}}$           & Current of a single DAC (for calibration)                & Current       & Analog          & BNC           & -             \\
$\mathrm{Ca^{2+}_{ABOVE}}$ & $\mathrm{Ca}^{2+}$ above high threshold                  & Voltage       & Digital         & Pin           & NRN          \\
$\mathrm{Ca^{2+}_{BELOW}}$ & $\mathrm{Ca}^{2+}$ below low threshold                  & Voltage       & Digital         & Pin           & NRN          \\
$\mathrm{POST_{ABOVE}}$    & Post trace above high threshold                          & Voltage       & Digital         & Pin           & NRN          \\
$W_{\text{SYN}}$           & Digitized $V_{\text{W}}$                               & Voltage       & Digital         & Pin           & SYN          \\
\textbf{Device Monitoring }         &                                                          &               &                 &               &              \\
$I_{\text{DEV-NEG}}$     & Current from negative device                             & Current       & Frequency             & sADC          & SYN          \\
$I_{\text{DEV-NORM}}$    & Current from normalizer circuit                                       & Current       & Frequency             & sADC          & SYN          \\
$\mathrm{DEV_{READ}}$    & Device read pulse                                    & Voltage       & Digital         & Pin           & SYN          \\
$\mathrm{DEV_{WRITE}}$   & Device write pulse                                   & Voltage       & Digital         & Pin           & SYN          \\
$\mathrm{DEV_{INT}}$     & Device interrupt flag                                & Voltage       & Digital         & Pin           & SYN          \\
$\mathrm{DEV_{STATE}}$   & Device state                               & Voltage       & Digital         & Pin           & SYN          
\end{tblr}
\vspace{10pt}
\caption{\textbf{Signals that can be monitored on the TEXEL chip.} 
The table is divided into three blocks: current signals observable through the spiking output of the \glspl{sadc}; voltage and current outputs measurable through \gls{bnc} connectors; and digital flags measurable on output pins.}
\label{tab:monitoring}
\end{table}

\subsection{sADC}
\label{appendix:sadc}
\begin{figure}
    \centering
    \begin{subfigure}[b]{0.45\textwidth}
        \centering
        \includegraphics[width=\textwidth]{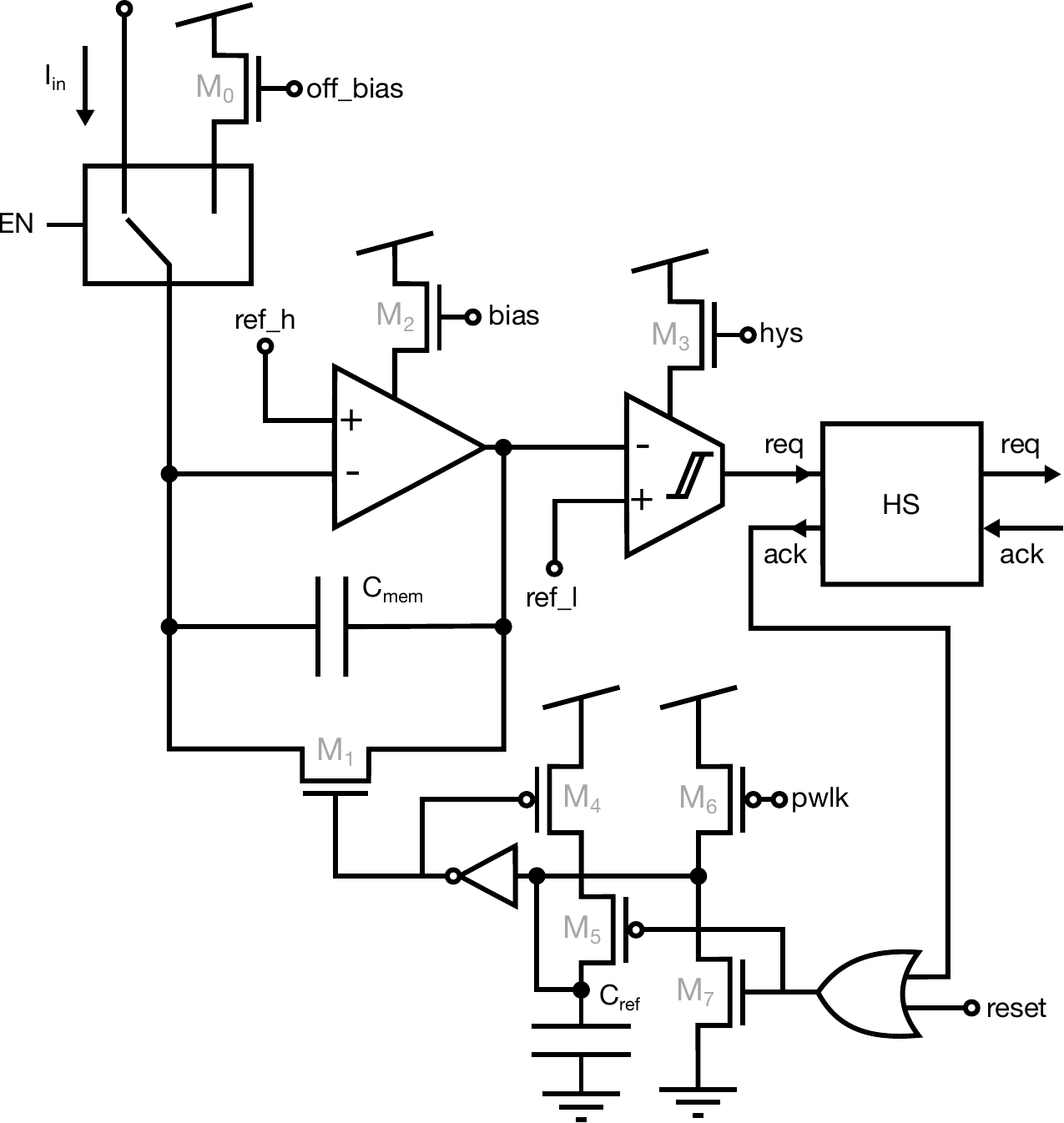}
        \caption{sADC schematic}
        \label{fig:sadc_schematics}
    \end{subfigure}
    \hfill
    \begin{subfigure}[b]{0.45\textwidth}
        \centering
        \includegraphics[width=\textwidth]{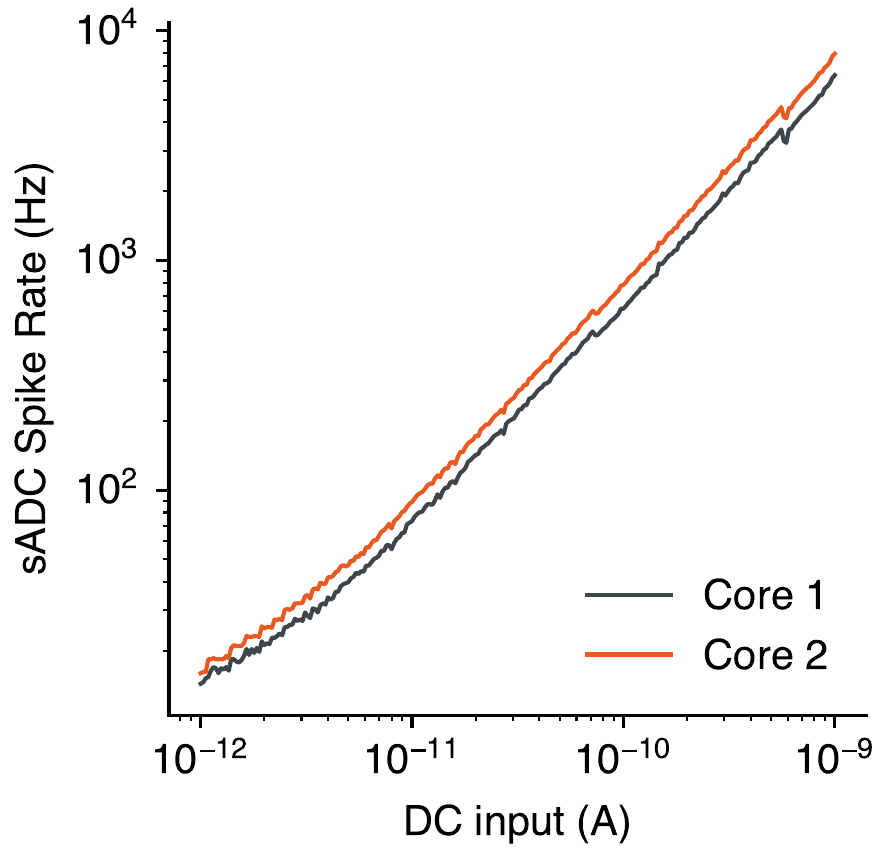}
        \caption{sADC response}
        \label{fig:sadc_response}
    \end{subfigure}
    \vspace{10pt}
    \caption{\textbf{\gls{sadc} circuit and the measurements of the spike rate in response to \gls{dc} input.} \textbf{a)} \gls{sadc} schematic with biases labelled. The input current is converted to a spikes and is transmitted via the handshake (HS) block and encoded as an address representing the signal being monitored by the \gls{sadc}. \textbf{b)} Spiking response of two \gls{sadc} circuits (one on each core) in response to a logarithmic sweep of DC input current from an on-chip \gls{dac}. The mapping between current and spike rate obeys a power law, these measurements show how the spike rate of an \gls{sadc} can be used to infer the magnitude of its input current. The core-to-core deviation is due to mismatch.}
\end{figure}

On the TEXEL chip 49682 currents can be monitored, of which 24 simultaneously. 
This is possible thanks to the implementation of the \gls{sadc}~\cite{Voulgari_etal15}. 
The circuit follows a mixed signal approach, where the analog block continuously interacts with the asynchronous digital block in a way inspired by mixed-signal implementations of spiking neurons. 
The working principle of the \gls{sadc} is as follows: the current under monitoring is mirrored from the circuit and fed in the input of the \gls{sadc}. 
This current is directed towards the negative input node of a \gls{opamp}, connected through a capacitor $\mathrm{C_{mem}}$, to the \gls{opamp}'s output, generating a negative feedback loop. 
The negative feedback loop, under ideal conditions, allows for the creation of a virtual ground: the negative input node of the \gls{opamp} stabilizes its voltage close to $\mathrm{ref\_h}$, regardless of the input current received, while allowing the input current to charge $\mathrm{C_{mem}}$. 
Charging the capacitor with said current, while the transistor gate ($\text{M}_1$) stays at a fixed voltage, results in an increasing output voltage, which is sensed by a the subsequent circuit. 
This circuit is composed of a hysteresis-equipped \gls{ota} which implements a threshold function. 
Here, the input voltage is compared to a fixed bias, and only when the input is above a certain voltage $\mathrm{CF_{REF-L}}+V_{\mathrm{HYS}}$, the output changes its digital state. 
The switch of the output state activates the asynchronous digital interface (HS), generating a request for a spike event. 
Once the circuit receives the acknowledgement signal from the subsequent digital block, the reset of the integrated current begins: a digital pulse completely discharges capacitor $\mathrm{C_{refr}}$, which is then promptly charged back by a constant current $\mathrm{CF_{PWLK}}$. 
The time taken by this capacitor to be charged sets the refractory period of the circuit. 
During this time, in fact, the capacitor $\mathrm{C_{mem}}$ has its terminals shorted by an active transistor ($\text{M}_1$), inhibiting the ability to charge with input currents. 
When no acknowledgement is detected, a pull-up is actively keeping the capacitor $\mathrm{C_{refr}}$ charged. 
The behaviour of the \gls{sadc} versus an input current can be seen in Figure~\ref{fig:sadc_response}, where, using a programmable \gls{dac}, the input current has been swept logarithmically between \qty{1}{\pico\ampere} to \qty{1}{\nano\ampere}. 
One can notice the very wide range of frequency at the output that demonstrates the ability of the circuit to monitor a very wide range of input currents.

The tuning parameters for the circuit are:

\begin{itemize}
    \item $\mathrm{EN}$: a digital flag determining whether the \gls{sadc} should receive inputs from the monitored signals or from $\mathrm{off\_bias}$.
    \item $\mathrm{off\_bias}$, a fixed bias current alternative to the input current.
    \item $\mathrm{ref\_h}$: The voltage at which the virtual ground should be set. This voltage shifts the low node of the capacitor $\mathrm{C_{mem}}$.
    \item $\mathrm{bias}$: the current at which the \gls{opamp} should be biased: it defines the strength of the feedback loop (so the ability of the circuit to keep the virtual ground to a specific voltage regardless of the input current magnitude and speed).
    \item $\mathrm{ref\_l}$: The voltage at which the capacitor's positive node is compared in the \gls{ota}.
    \item $\mathrm{hys}$: the current deciding the hysteresis value of the \gls{ota}. This defines how much capacitor positive node should be offset with respect to $\mathrm{ref\_l}$, to elicit a spike, such that $\mathrm{V_{mem} > ref\_l+V_{hys}}$.
    \item $\mathrm{pwlk}$: the leakage of the refractory transistor, this parameter sets how long should the circuit wait before being able to integrate current again.
\end{itemize}

\subsection{DAC}
\label{appendix:dac}

Each core incorporates a fully programmable 94-channel 12-bit \gls{dac}, capable of generating reference currents and parameters ranging from \qty{0.5}{\pico\ampere} to \qty{2.2}{\micro\ampere}, inspired by the design proposed in~\cite{Delbruck_Schaik05}. These parameters serve to configure the operation settings for the neuron, synapse and learning circuits, as well as control the timing of the device interface. 
The \gls{dac} is comprised of three components: first, the configuration storage, which is part of the digital blocks. 
Second, the reference current generator, responsible for producing six reference currents. 
Third, the 1T-2T current dividers to generate channel currents from these references. 
The reference current segment employs a subthreshold CMOS and resistor \gls{ptat} source, augmented by a current divider block that incorporates a resistor in the divider to function as a \gls{ctat} source. 
Together with the \gls{ptat}, this combination reduces the temperature sensitivity. 
The resulting current is directed into scaling current mirrors and conveyors to generate scaled base currents (master currents). 
In this instance, master currents include values of \qty{2.2}{\micro\ampere},
\qty{0.29}{\micro\ampere},
\qty{36}{\nano\ampere},
\qty{4.5}{\nano\ampere},
\qty{0.57}{\nano\ampere},
and \qty{70}{\pico\ampere}.
For each channel, one of these currents is chosen. 
The resulting current is then passed to a finer division stage of 8 bits, providing 256 levels, with the last selection per channel determining whether the current is sourced by an nFET or a pFET. 
The fine division stage is composed of MOSFETs substituting resistors in the common 1R-2R \gls{dac} circuit (here called 1T-2T). 
Due to the fact that the current divider employs MOSFETs instead of resistors, the saturation condition of the transistors need to be guarded. 
The result of the violation of the saturation condition of the MOSFET is the \gls{dac} not being entirely monotonic.

\subsection{Device Operation \& Integration}
\label{appendix:dev_operation}

\subsubsection{Synapse Controller}

\begin{figure}
    \centering
    \includegraphics[width=\textwidth]{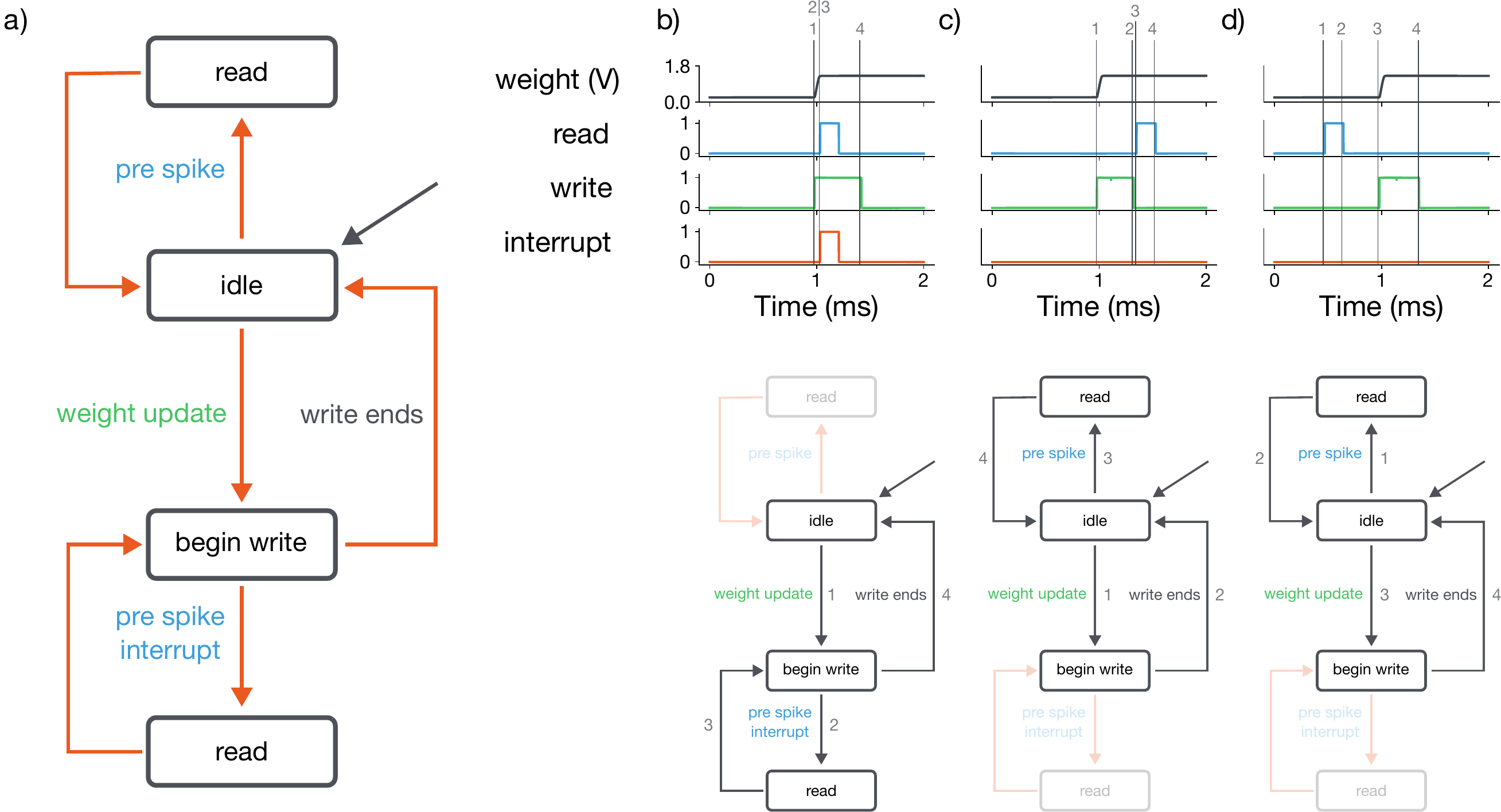}
    \vspace{5pt}
    \caption{\textbf{Device synapse controller state machine and silicon measurements of the digital flags.} 
    \textbf{a)} A state diagram delineating the internal states of the digital logic governing the addressing protocol for devices on TEXEL. 
    Each row represents a potential combination of presynaptic spikes, postsynaptic spikes, and the state update of synaptic weights. 
    When a presynaptic spike is present, the synaptic weight is read. 
    If a weight update is triggered by either a pre or postsynaptic spike, the logic initiates the writing protocol for the differential device setup. 
    The sequence of these events is unproblematic unless a write and read request are concurrently issued.
    In such a scenario, a read request takes precedence, and any write request is temporarily halted to allow for the read to take place. 
    Following the reading of the devices, the write process is subsequently executed. 
    \textbf{b)} Silicon measurements of digital flags raised by the device controller circuitry located within each synapse. 
    A device read occurs at the same time as a device read, in this case an interrupt flag is raised such that a read can be prioritised and write is subsequently executed. 
    \textbf{c)} A write occurs, due to a synaptic weight change, and a read follows. 
    No interrupt flag is raised. 
    \textbf{d)} A read, due to a presynaptic spike, occurs before a write. No interrupt flag is raised.}
    \label{fig:dev_logic}
\end{figure}

We conducted chip measurements to verify the functionality of the synapse controller. 
The synapse controller constitutes circuitry at each synapse which manages cases where read and write operations overlap.
Three scenarios of read-write interactions are examined through read and write protocols, with digital pins capturing read, write, and interrupt pulses, and weight changes monitored via analog channels. 
The controller appropriately prioritizes read operations over writes, as evidenced by the detection of interrupt flags when a read coincides with a write pulse, and then executes the write; this ensures correct device operation (Fig.~\ref{fig:dev_logic}). 

\subsubsection{Continuous Read}

In addition to the aforementioned device operation mode, the TEXEL platform can be configured to a ``continuous read'' mode. 
In this mode the READ signal is permanently set to high such that the drain of the device is held at $V_\text{read,D}$, the gate is held at $V_\text{read,G}$ and the source is connected to both input branches of the normalizer circuit (Fig.~\ref{fig:device_interface}). 
This READ state is mutually exclusive with respect to memristive device writing signals ($\mathrm{POT}$/$\mathrm{DEP}$). In this mode the $\mathrm{IDLE}$ signal becomes obsolete and is held at ground.
This ``continuous read"" mode would be used in the case for which the memristive device capacitance is high and potentially outside ``compatibility'' range derived from simulations.

\subsubsection{Transition-Metal Oxides}

Two-terminal memristive devices consisting of one or more layers of transition metal oxides are widely used for neuromorphic systems, especially for emulating synaptic functions~\cite{Ziegler_etal18}. 
Here we evaluate three-layer memristive device stacks for their integration into the TEXEL platform. 
The layer sequence of the memristive device considered for this purpose is HfO$_x$ /Al$_2$O$_3$ /TiO$_2$, embedded between an Au contact layer and the TiN bottom electrode~\cite{Park_etal22}. 
Here, the HfO$_x$ is responsible for the memristive behavior, the Al$_2$O$_3$ changes the interface properties and the TiO$_2$ layer is advantageous because it forms well-defined interfaces with the TiN electrode and the Al$_2$O$_3$ intermediate layer. 
Furthermore, the Al$_2$O$_3$ layer controls the generation of oxygen vacancies and thus serves to limit the current. 
This is particularly important for integrating the devices into circuits in order to operate the devices without a current compliance. 

The stoichiometry of the HfO$_x$ layer is decisive for the resistive switching mechanism~\cite{Park_etal22}. 
Particularly for sub-stoichiometric oxides ($x$ between \num{1.5} and \num{1.8}), filamentary switching is observed, while stoichiometric oxide layers ($x = 2$) have an interface-based switching mechanism. 
However, this leads to different device properties. Gradual resistance switching is observed in interface switching devices, while devices based on filamentary switching exhibit a more abrupt switching characteristic. 
In the latter, however, multi-level resistance states can be achieved by careful design of the oxygen-vacancy filament. However, the two classes of devices have different requirements that need to be considered when integrating them into the TEXEL platform, which we have analyzed below. Both types of switching devices were fabricated in a thin-film technology using reactive DC magnetron sputtering. 
This was used to deposit the layers of the device stack with the following thicknesses: HfO$_x$  has a thickness of \qty{3}{\nano\meter}, Al$_2$O$_3$ of \qty{2}{\nano\meter} and TiO$_2$ of \qty{15}{\nano\meter}. 
The device electrodes are electrically insulated by a \qty{180}{\nano\meter} thick SiO$_2$ layer, encapsulating the functional layers. 
A \qty{30}{\nano\meter} thick Au layer defines the top electrode and are used to define the active device area. 
Further details on the device fabrication can be found in~\cite{Park_etal22}.  

\textbf{Interface switching devices:} 
The resistance values for this class of devices are between \qty{0.7}{\mega\ohm} and \qty{290}{\mega\ohm}, depending on the concentration of oxygen vacancies in the active memristive HfOx layer. 
Here, $R_{\mathrm{on}}/R_{\mathrm{off}}$ ratios of up to \num{e-3} are achieved. 
The switching voltages required for this are \qty{2.5}{\volt} or \qty{3.5}{\volt} for setting the devices and \qty{-1.5}{\volt} or \qty{-3}{\volt} for resetting. 
This corresponds to current values of \qty{3.6}{\micro\ampere} and \qty{10}{\nano\ampere} as well as \qty{-2.1}{\micro\ampere} and \qty{-10}{\nano\ampere}. 
In other words, values that are compatible with the TEXEL platform (Fig.~\ref{fig:device_compatibility}c). 
However, these values are dependent on the device area and were determined for an area of \qty{20}{\micro\meter\squared}. 
For a direct integration of these devices on the contact areas shown in Fig.~\ref{fig:device_interface}a, a reduction of the device area by a factor of about 10 is necessary. 
However, this would be accompanied by a moderate increase in resistance. 
This can be estimated from a resistance value in the off state of \qty{20}{\mega\ohm} for the current area size to \qty{40}{\mega\ohm} - \qty{50}{\mega\ohm} if the device area size is reduced by a factor of \num{10}. 
Values that the TEXEL platform allows. 

Another important device parameter that must be determined and adapted for the integration of the devices into the TEXEL platform is the device/layer capacitance. 
The layer capacitance of the transition metal oxides in the layer thickness range used here can be estimated in the range of \qty{e-14}{\farad\per\micro\meter\squared}~\cite{Hansen_etal15}, which fulfils the requirements of the TEXEL platform given in Fig.~\ref{fig:device_compatibility}d.

\textbf{Filamentary switching devices:} For filamentary memristive devices, the resistance values of the off resistance are in the range of \qty{0.16}{\mega\ohm} - \qty{80}{\mega\ohm} depending on the concentration of oxygen vacancies in the HfO$_x$ layer~\cite{Park_etal22}. 
Voltages from \qtyrange{-1.5}{-3.0}{\volt} are required for setting the devices, while resetting requires voltages in the range \qtyrange{2.5}{3.0}{\volt}. 
This can be converted into current values in the range \qty{-20}{\micro\ampere} to \qty{-20}{\nano\ampere} for the setting process and \qty{15}{\micro\ampere} to \qty{40}{\nano\ampere} for the resetting process. 
The $R_{\text{on}}/R_{\text{off}}$ ratio with is \numrange{e-1}{e-2}, slightly smaller compared to interface switching devices, but fulfils the requirements of the TEXEL platform very well, as shown in Fig.~\ref{fig:device_compatibility}c. 
Capacitance is determined by the device area as well as the filament area. 
The relevant size for integration is the device area capacitance, which we assume to be \qty{e-14}{\farad\per\micro\meter\squared} as in the case of interface switching devices. 
The scaling of the device area has no influence on the resistance values. 
However, an inertial forming step is required for these devices, which requires voltages of up to \qty{\pm5}{\volt}, which is compatible with the TEXEL platform. 

\subsubsection{Ferroelectric Hafnia}

Two- and three-terminal synaptic weights based on ferroelectric hafnia are evaluated for their integration on the TEXEL platform. 
In the two-terminal configuration, the current flows through the ferroelectric layer, which requires the layer thickness to be scaled while maintaining a high polarization. 
The materials were specifically developed on the XFAB \qty{180}{\nano\m} technology, replicating the conditions for integration on the TEXEL platform. 
The conductivity and the dynamic range of the \gls{beol}-integrated synaptic weights were found to differ from the same weights nanofabricated on dummy Si wafers~\cite{begon2024back}. 
The \qty{2}{V} required to operate the synaptic weights fall well in the range available in TEXEL. 
The dynamic range falls between \num{1} and \num{10}, for which the optimal resistance is predicted to be in the \qty{10}{\giga\ohm} range. 
It would result in an average current sourced by the differential normalizer synapse of two thirds of $\mathrm{norm\_bias}$. 
The scalability of the resistance with the area allows to adapt the design to the current requirement: an ideal resistance of \qty{10}{\giga\ohm} is obtained by scaling the device to \qty{10}{\micro\meter\squared}.

In the three-terminal configuration (\gls{fefet} or thin-film transistor) the ferroelectric gate is integrated prior to the semiconducting oxide channel. 
The materials optimized for the fabrication of two-terminal devices on TEXEL were evaluated for three-terminal devices, i.e. with an increased gate thickness up to \qty{10}{\nano\meter}. 
In test circuits, the ferroelectric switching of the capacitors was demonstrated through the same interconnects and transistors that on the TEXEL platform~\cite{hamming2024multi}. 
The saturation for the ferroelectric switching is obtained for \qty{\pm4}{\volt}, in line with the device requirements.

The CMOS-compatibility translates in the absence of degradation of the front-end electronics during the back-end integration of the synaptic weights. 
For the ferroelectric technology presented above, the critical steps are:

\begin{enumerate}
    \item the deposition of a functional tungsten oxide layer at \qty{375}{\degreeCelsius} under an oxidizing plasma
    \item the crystallization of hafnia in the ferroelectric phase
\end{enumerate}

It uses a flash lamp annealer applying a \qty{20}{\milli\second} long energy pulse of \qty{90}{\joule\per\centi\meter\squared}, at a temperature of \qty{375}{\degreeCelsius}. 
The XFAB \qty{180}{\nano\meter} MOSFET characteristics prior and after the ferroelectric device integration were compared and did not show significant changes~\cite{begon2024back}. 
These preliminary results represent a first milestone towards the evaluation of two- and three-terminal synaptic weights based on ferroelectric hafnia using the TEXEL neuromorphic processor.

\renewcommand{\arraystretch}{1.5}

\begin{figure*}
    \centering
    \begin{subfigure}[b]{0.6\textwidth}
        \centering
        \includegraphics[width=\textwidth]{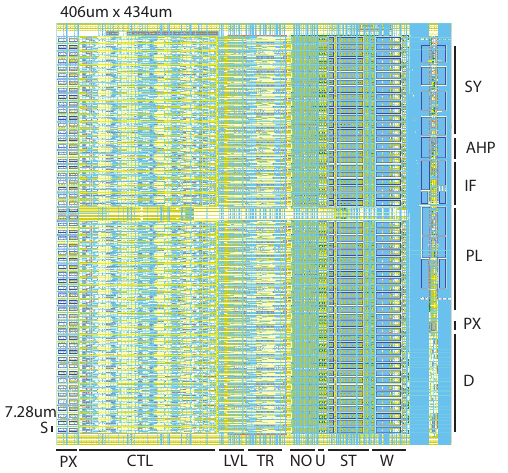}
        \caption{Neuron block footprint}
    \end{subfigure}
    \hfill
    \begin{subfigure}[b]{0.39\textwidth}
        \centering
        \resizebox{\linewidth}{!}{
        \begin{tabular}{lll}
            \toprule
            \textbf{Abbreviation} & \textbf{Name}                         & \textbf{Area \%}                                  \\
            \midrule
            \multicolumn{3}{c}{\textbf{Synapse}}                                                             \\ \midrule
            S            & Synapse                      & -                                         \\ 
            PX           & Pulse Extender               & 5.4                                       \\
            CTL          & Device Controller            & 31.5                                      \\ 
            LVL          & 1.8-5V Level Shifters      & 5.8                                       \\ 
            TR           & Device Interface             & 9.9                                       \\ 
            NO           & Normalizer                   & 6                                         \\ 
            U            & Post Synaptic Current        & 2.2                                       \\ 
            ST           & Synaptic Trace               & 10.6                                      \\ 
            W            & Bistable Weight/Update Logic & 4                                         \\ \midrule
            \multicolumn{3}{c}{\textbf{Soma}}                                                              \\ \midrule
            SY           & PSC Filter                   & 0.4                                       \\ 
            AHP          & Spike Frequency Adaptation   & \multicolumn{1}{l}{\multirow{2}{*}{0.34}} \\
            IF           & Leaky Integrate and Fire     & \multicolumn{1}{l}{}                      \\ 
            PL           & Post Learning Traces         & 0.43                                      \\ 
            D            & Neuron Digital Logic         & 0.41                                      \\ 
            & {Wiring and Power}        & 23.02                                     \\ \bottomrule
        \end{tabular}}
        \vspace{10pt}
        \caption{Key}
        \label{fig:neuron_block_key}
    \end{subfigure}
    \vspace{10pt}
    \caption{\textbf{The neuron block macro of TEXEL, detailing the location and footprint area of the circuits.} \textbf{a)} 
    The footprint of the neuron block with associated labels and sizings. \textbf{b)} Table defining the abbreviations and providing the \% of area of the neuron block macro they occupy.}
    \label{fig:neuron_block}
\end{figure*}

\begin{figure*}
    \centering
    \includegraphics[width=0.75\linewidth]{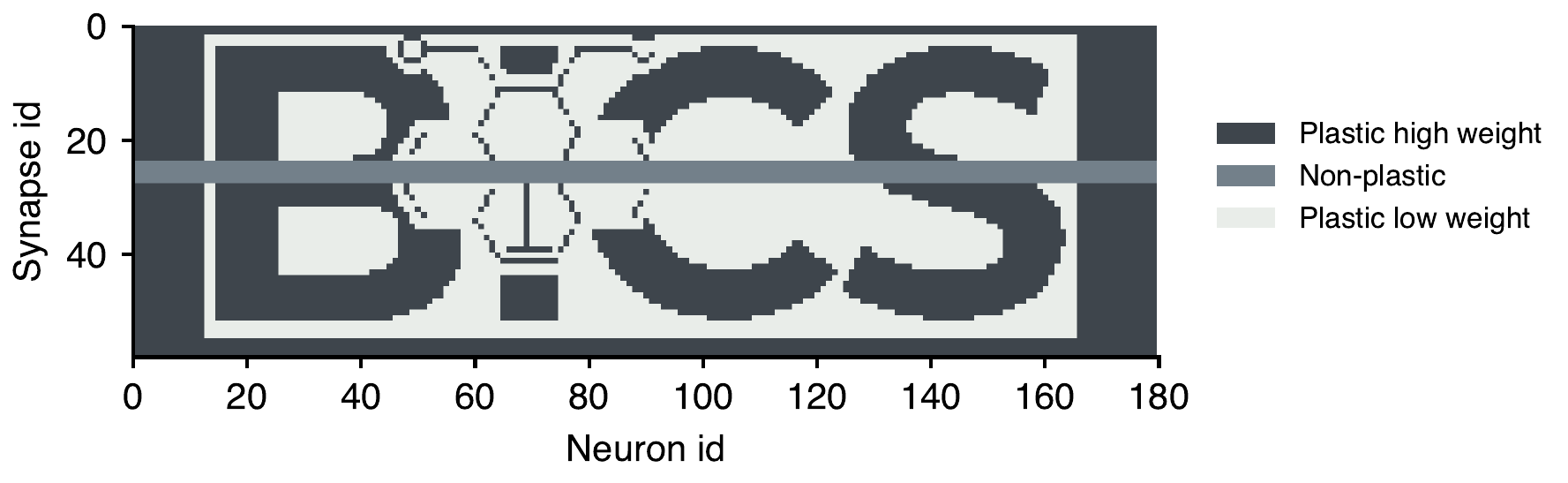}
    \caption{\textbf{Readout of a weights matrix programmed onto the plastic synapses of TEXEL.} 
    Each of the 54 plastic synapses has the capability to exist in either a high or low state, effectively storing binary information. 
    A weights matrix can be programmed onto the chip, making it suitable for inference tasks, device operation, and testing; independent of on-chip learning.}
    \label{fig:inference}
\end{figure*}

\clearpage
\newpage

\printbibliography[heading=subbibliography,title={Supplementary References}]
\end{refsection}

\end{document}